%% file: graph_representation.tex
\documentclass[journal]{IEEEtran}
\usepackage{amsmath}
\usepackage{algorithm}
\usepackage{graphicx}
\usepackage[noend]{algpseudocode}
\usepackage{amsthm}
\usepackage{enumitem}
\usepackage{caption}
\usepackage{subcaption}
\usepackage{mathtools}
\usepackage{tikz,listings}
\usepackage{array}
\usepackage{enumitem}
\usepackage{multicol}

\newtheorem{observation}{\textbf{Observation}}
\newtheorem{corollary}{\textbf{Corollary}}

\newtheorem{Constraint}{\textbf{Constraint}}
\usetikzlibrary{decorations.pathreplacing,calc}
\newcommand{\tikzmark}[1]{\tikz[overlay,remember picture] \node (#1) {};}

\newcommand*{\AddNote}[4]{%
    \begin{tikzpicture}[overlay, remember picture]
        \draw [decoration={brace,amplitude=0.3em},decorate,thick,black]
            ($(#3)!([yshift=1.5ex]#1)!($(#3)-(0,1)$)$) --  
            ($(#3)!(#2)!($(#3)-(0,1)$)$)
                node [align=center, text width=1cm, pos=0.5, anchor=west] {#4};
    \end{tikzpicture}
}%
\DeclarePairedDelimiter{\ceil}{\lceil}{\rceil}
\DeclareRobustCommand{\Leadsto}{%
  \mathrel{\text{\usefont{U}{lasy}{m}{n}\symbol{"3B}}}%
}

\title{A Prufer-Sequence Based Representation of Large Graphs for
Structural Encoding of Logic Networks}

\author{ Manjari Pradhan and Bhargab B. Bhattacharya ~\textit{Fellow, IEEE}
			}

\begin{document}
\maketitle

\begin{abstract}

The  pervasiveness of graphs  in today's real life systems is quite evident, where the system  either  explicitly exists as graph  or can be readily modelled as one. Such graphical structure is thus a store house rich information. This has various implication depending on whether we are interested in  a node  or the graph as a whole. In this paper, we are primarily concerned with the later, that is, the inference that  the structure of the graph influences the property of the real life system it represents.  A model of  such structural influence would be useful in inferencing useful properties of complex and large systems, like VLSI circuits,  through its structural property. However, before we can apply some machine learning (ML) based technique to model such relationship, an effective representation of the graph is imperative. In this paper, we propose a graph representation which is lossless, linear-sized in terms of number of vertices and gives a 1-D representation of the graph. Our representation is based on Prufer encoding for trees. Moreover, our method is based on a novel technique, called $\mathcal{GT}$-enhancement whereby we first transform the graph such that it can be represented by a  singular tree. The encoding also provides scope to include additional graph property and improve the interpretability of the code.
\end{abstract}


\section{Introduction}
\label{sec:introduction}
Graph-based machine learning have seen a surge of research, recently, driven by the prevalence of graphs in the real world systems.  
 In such environments, either the data naturally  occur as graphs like in biological networks, or the  interactions are modeled through graphs as in social networks.  The graph based ML have been applied to varied  of problem setting with a range of unique approaches \cite{ISARX20}.  While ML has gained stupendous proficiency in its tools and techniques as well as huge success in its application, its downside lies in the fact that the basic mathematical operations, that it relies on, necessitates the data to be in a structured format, which may be a feature vector or and an image grid. However, a graphical data is inherently not structured. So, most of the graph-based ML approaches aim at capturing maximum relevant information in a structured format.
 We next discuss the present works and the different problem setting and approaches.

The real world environment which are represented by graph vary widely in the nature of the graph, the kind of interactions and also their sizes from large graph representing systems like VLSI circuits to extra large like those representing social network. Hence these factors determine the problem setting and approaches. The approach taken also  is determined by the precision demanded in terms of the features for ML2  and the scope  of representing the graph which is constrained by its size . Although there are many problem setting looked at so far   \cite{ISARX20}, the two generic problem setting has evolved around whether we are interested in obtaining the information about a node or the graph as a whole. While most of the research is focused on the former, the later  has fewer solutions in literature. One of the reasons for this is it is challenging to represent the entire graph because of its complexity as well as its size.  A number of surveys are available in the literature which focus on three basic approaches: (i) graph embedding \cite{HVTKDE18,FMSPC12}, (ii) representation learning \cite{LWCorr17}, and (iii) geometric deep learning \cite{MLSPM17}. Graph embedding aims at representing the graph as a vector \cite{FMSPC12} or as a set of vectors \cite{HVTKDE18}. The properties are  based on local proximity, which are either manually defined or  extracted in a automated fashion. The latter  belongs to the class of representation learning frameworks where feature formulation is automated using ML or matrix factorization  \cite{LWCorr17}. Geometric deep-learning tools for graphs are aim at applying deep learning tools like convolutional neural nets (CNN) or recurrent neural network (RNN). Vector representations have mostly been used in a node-centric setting where  a node is represented by a low-dimensional vector. They are mostly used for node classification or link prediction \cite{GKDD16,LINE15}. Geometric deep learning are based on one of the two techniques, spectral \cite{DBNIPS16} or spatial approach \cite{MMICML16}. Spectral analysis have the following major bottlenecks \cite{MMICML16}: (i) they are valid only for undirected graphs, (ii) applicable only to graphs with similar size and structure, and (iii) learning is mostly based on the weights of the vertices of the graphs.  A  few works aim at capturing full graph embedding/learning, e.g., graph kernel or graph neural network (GNN) \cite{FMTNN09}. However, none of these preserve the graph structure,  or are  applicable to large graphs. Although CNNs based on spatial analysis have been used to handle arbitrary graphs, complete structure preservation is still a concern  \cite{MMICML16}.

 Recently, in the field of integrated circuits, various methods have been employed for efficient application of ML and deep learning tools \cite{GWCoRR17,WEISCAS18,YRHOST17,YHDAC19,MBWIRES20}. One of the rich source of graph based data is the graphical structure of the logical circuits called circuit-graphs.  Circuit-graphs are directed acyclic graphs (DAG) representing combinational or scan-based logic circuits.  It has been seen that the structural features of the circuit-graph can be used to infer the some property of the circuit \cite{MBTCAD18}.  Since devising the structural features in laborious some ML based techniques to learn the structural features would expedite the process.  However, such type of graph based learning has not been addressed in any of the previous methods. A major requirement here is the need for an effective graph representation for which there is no proper existing method.

In this paper, we propose a representation of graphs, which are large, and which fulfill the following properties that are required for the afore-mentioned problem setting. Firstly, it is  lossless. The structure of the graph is preserved in the proposed representation. Secondly, it is linear sized in terms of the number of vertices. Hence it is convenient to store large sparse graphs. Thirdly, we propose a 1D representation of graph. Such a representation, we believe would be convenient for ML based applications since most of the ML tools work on vectors of 2-D grid data.

We introduce a novel  encoding technique for this purpose. The encoding is lossless and is based on a very old but not fully explored, graph theoretic concept  known as Pr{\"u}fer sequence \cite{PAMP18}. This sequence was first used in 1918 to prove Cayley's formula, which was used to count the number of possible spanning trees in a graph with a given number of vertices. The classical Pr{\"u}fer code can be used for encoding trees only; however we are concerned with  graphs. For encoding of a graph, we need to  represent it with a tree. So, we propose a technique called \emph{graph-to-tree enhancement} (\emph{$\mathcal{GT}$-enhancement}), for this purpose. We call such a tree, which represents a graph through $\mathcal{GT}$-enhancement,  a  \emph{g-tree}. We present two approaches for $\mathcal{GT}$-enhancement. The first approach is based on DFS traversal and  relies on some additional operations to obtain a desirable \emph{g-tree}. In the second approach, we have simpled the method by introducing a novel graph traversal technique called SCESOR. the Further, we report new properties of Pr{\"u}fer codes and discuss methods for improving interpretability  and preserving the edge directions. We also discuss  Pr{\"u}fer codes in the light of making them learnable. 

The rest of the chapter is organized as follows.  The motivation and methodology appear in Section \ref{sec:motivation_ch5} and Section \ref{sec:methodology_ch5}, respectively. Section \ref{sec:gt-tree_based} and Section \ref{sec:gt-setsor} report the two methods that can be used to obtain a \emph{g-tree}. Section \ref{sec:code_selection} provides a discussion on the properties of Pr{\"u}fer code. Conclusions and future work appear in Section \ref{sec:conclusion_ch5}.

\section{Motivation}
\label{sec:motivation_ch5}
We will discuss the motivation highlighting three aspects. Firstly we will discuss  effectiveness of Pr{\"u}fer codes in representing large graphs representing real world systems. Secondly,  that although  prufer sequence is applied only to trees,  it can be easily applied to graphs by representing a graph as a single tree through the proposed methods of $\mathcal{GT}$-enhancement, which is discussed in later section. Thirdly, we give an illustrative example to demonstrate the capability of a Pr{\"u}fer sequence to preserve the entire structure of a circuit-graph such that the structure to the graph can be regenerated from the prufer code.

\begin{figure}[b]
\begin{subfigure}{0.6\textwidth}
\centering
\includegraphics[scale=0.35]{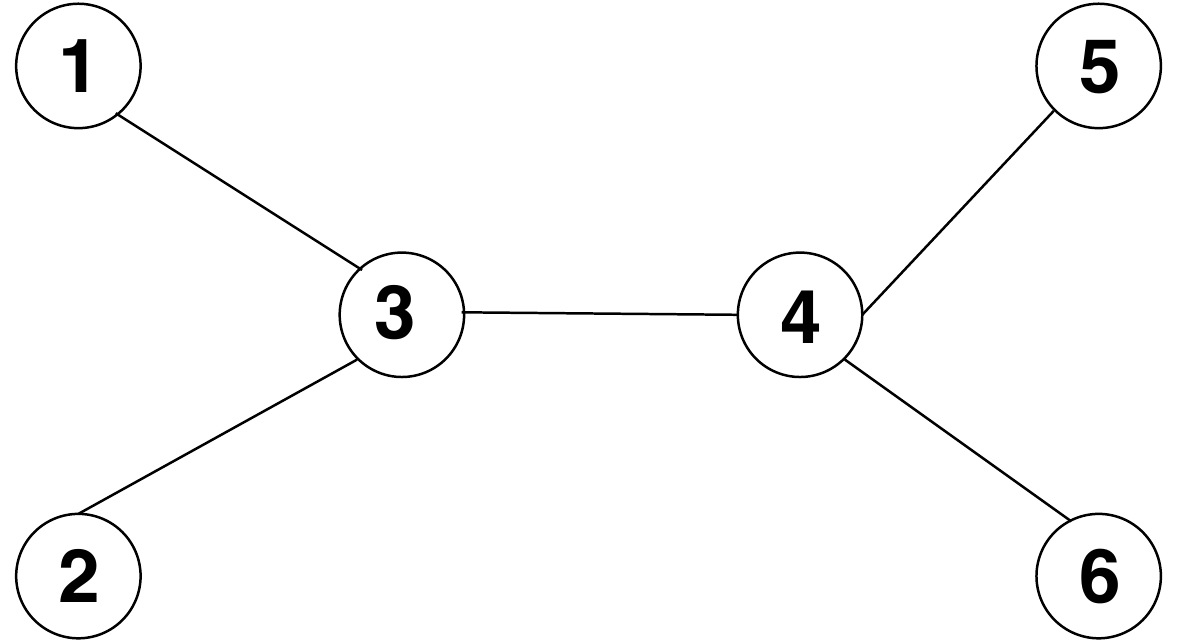}
\caption{Example labeled tree.}\label{fig:pruf_tree_eg_a}
\end{subfigure}
\begin{subfigure}{0.6\textwidth}
\centering
\vspace{1cm}
\includegraphics[scale=0.5]{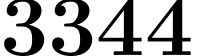}

\caption{Pr{\"u}fer code of the tree in Figure \ref{fig:pruf_tree_eg_a}}\label{fig:pruf_tree_eg_b}
\end{subfigure}
\caption{Example of Pr{\"u}fer code of a tree.}\label{fig:pruf_tree_eg}
\end{figure}

As mentioned before, the aim of this paper is to devise a representation of the graph which completely preserves the connectivity between the vertices. However,  this would be difficult for large graphs since it would require large memory. For example an adjacency matrix would require $n^2$ space for a graph of size $n$. Any effort to lessen the size would mean compromising on the connectivity information which we do not want. An interesting and favorable, observation about the graphs, that appear in real world, is that they are typically  sparse graphs. This implies that for a graph $G(V,E)$, where $V$ is the vertex set and $E$ is the edge set of $G$, $|E| \ll |V|^2$. In general, in sparse graphs  $|E| = \mathcal{O}(|V|)$. In fact, for most of the digital test benchmark circuits \cite{iscas89,itc99}, $|E|$ is not more than $2\times |V|$. Thus, a representation of the connections (edges), as a \textit{series of edges} would significantly shrink the size of the code.  A straight forward way of lossless representation is to consider all the possible vertex-pair and mark those which are connected as in done in adjacency matrix. But, here we want to keep only the marked vertex pairs (edges) and refrain  from retaining the rest of the vertex pairs in our code. The Pr{\"u}fer sequence provides a elegant method for such representation.
A Pr{\"u}fer encoding represents a labeled tree of $n$ nodes (and hence having $n-1$ edges) by a string of vertex-labels whose length is $n-2$.  Figure \ref{fig:pruf_tree_eg} shows an example of a Pr{\"u}fer code of a tree. A labeled tree with six vertices is shown in Figure \ref{fig:pruf_tree_eg_a} and its corresponding Pr{\"u}fer code is shown in Figure \ref{fig:pruf_tree_eg_b}.  The algorithm for encoding the tree and decoding it is discussed in the next section. We will also discuss how each vertex label represents an edge in later section.

\begin{figure}[h]
\begin{subfigure}{0.5\textwidth}
\centering
\includegraphics[scale=0.35]{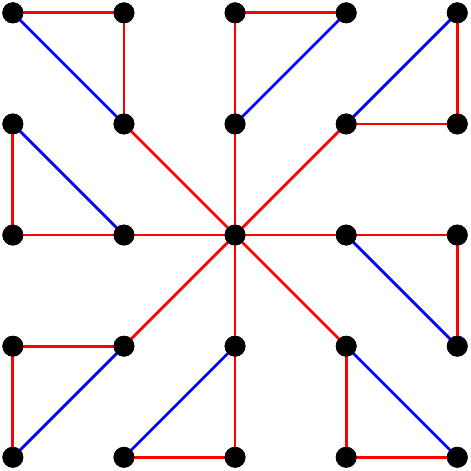}\hspace{2cm}\includegraphics[scale=0.35]{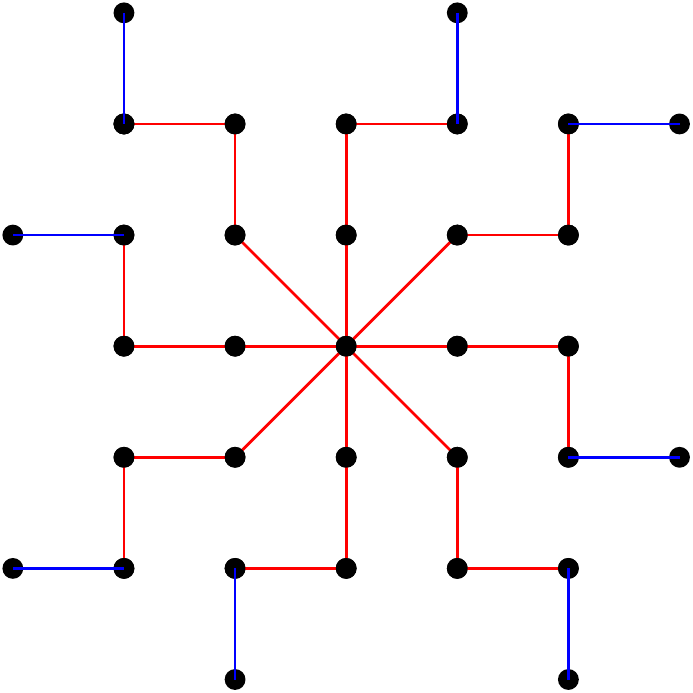}
\caption{}
\end{subfigure}
\begin{subfigure}{0.5\textwidth}
\vspace{1cm}
\centering
\includegraphics[scale=0.4]{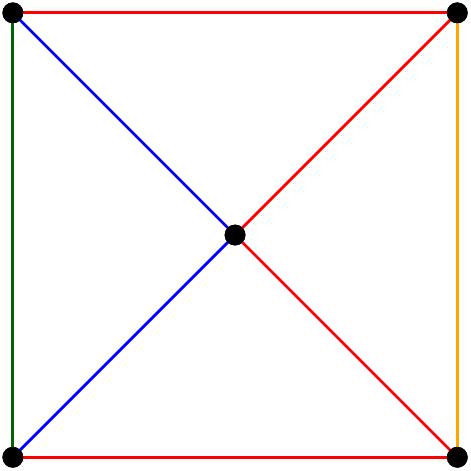}\hspace*{1cm}\includegraphics[scale=0.4]{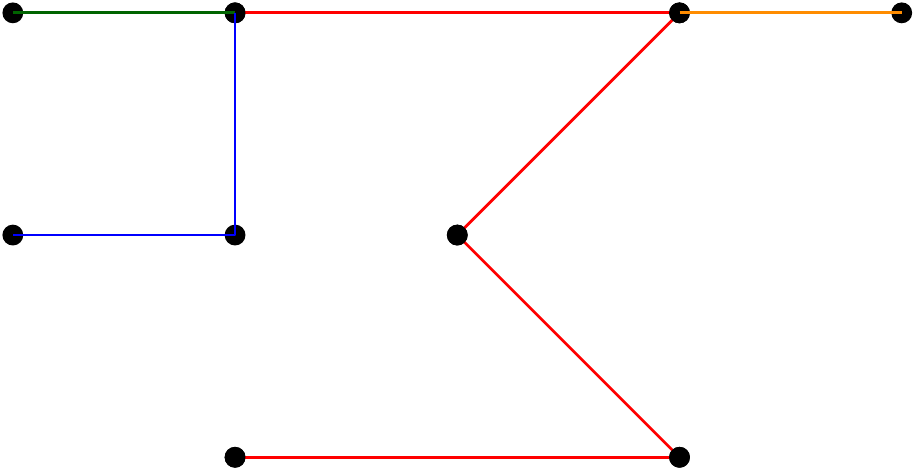}
\caption{}
\end{subfigure}
\caption{Graphs with large  number of tree-partitions. Top: (a) Sparse graph with  25 vertices 32 edges and 9  tree partitions ; (b) Dense graph with 5 vertices 8 edges and 4  tree partitions. Left: the corresponding \emph{g-trees}.}\label{fig:tree_partition}
\end{figure}
\begin{figure*}[!t]
\begin{tikzpicture}
\begin{scope}[xshift=1cm]
\node {\includegraphics[scale =0.43]{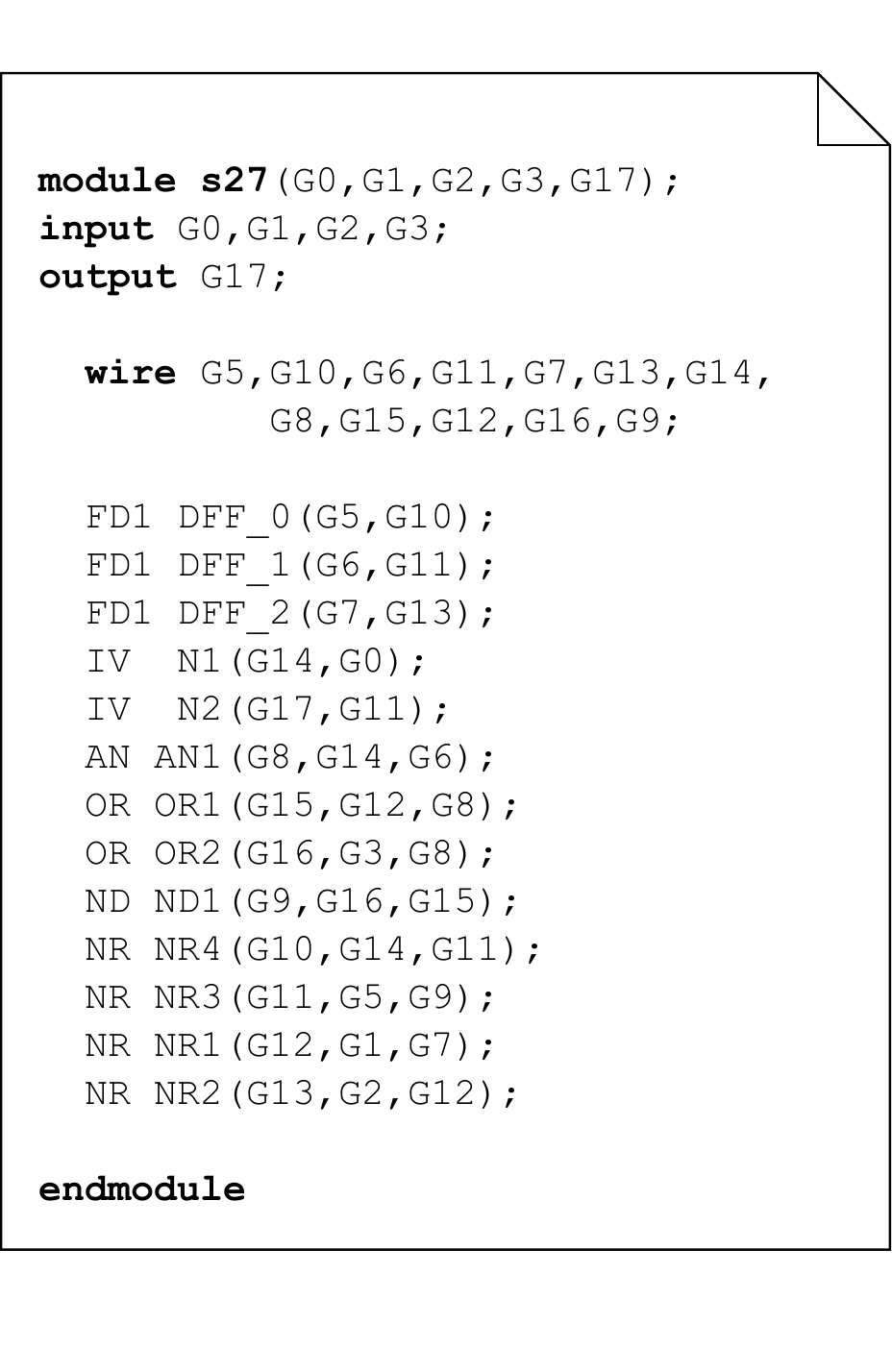}};
\end{scope}
\begin{scope}[xshift=1cm,yshift = -3cm]
\node {(a) netlist};
\end{scope}
\begin{scope}[xshift=6cm,yshift = 1cm]
\node {(b) circuit diagram};
\end{scope}
\begin{scope}[xshift=1cm,yshift = 3.5cm]
\node {\includegraphics[angle=180,scale =0.25]{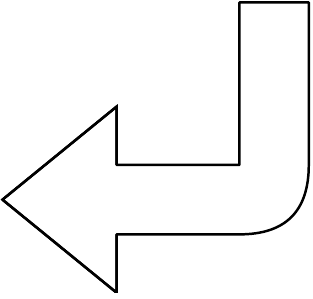}};
\end{scope}
\begin{scope}[xshift=12cm,yshift = 3cm]
\node {\includegraphics[angle=90,scale =0.25]{figures/s27/arrow_turn.pdf}};
\end{scope}
\begin{scope}[xshift=7cm,yshift = 4.2cm]
\node {\includegraphics[scale =0.3]{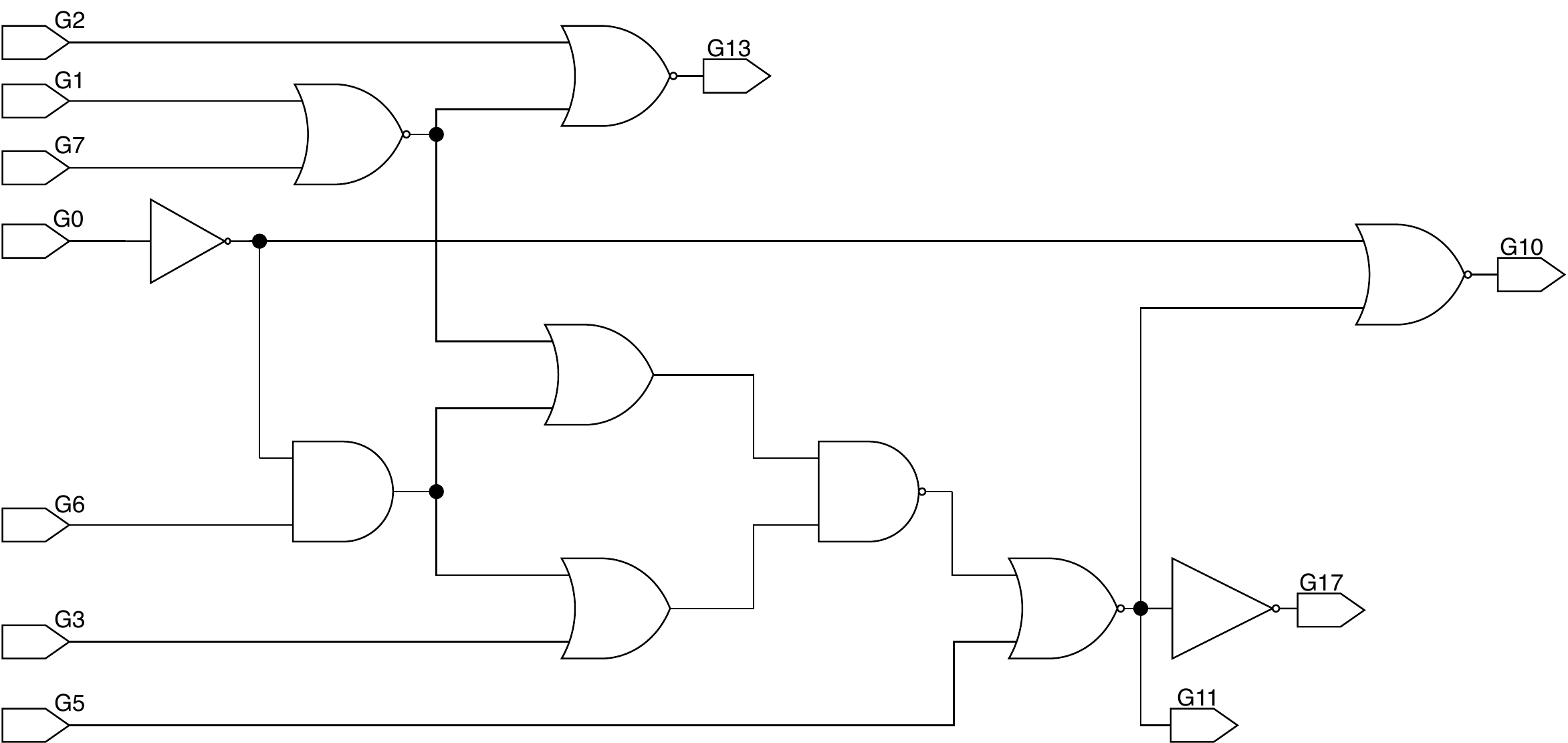}};
\end{scope}
\begin{scope}[xshift=11.5cm]
\node {\includegraphics[scale =0.48]{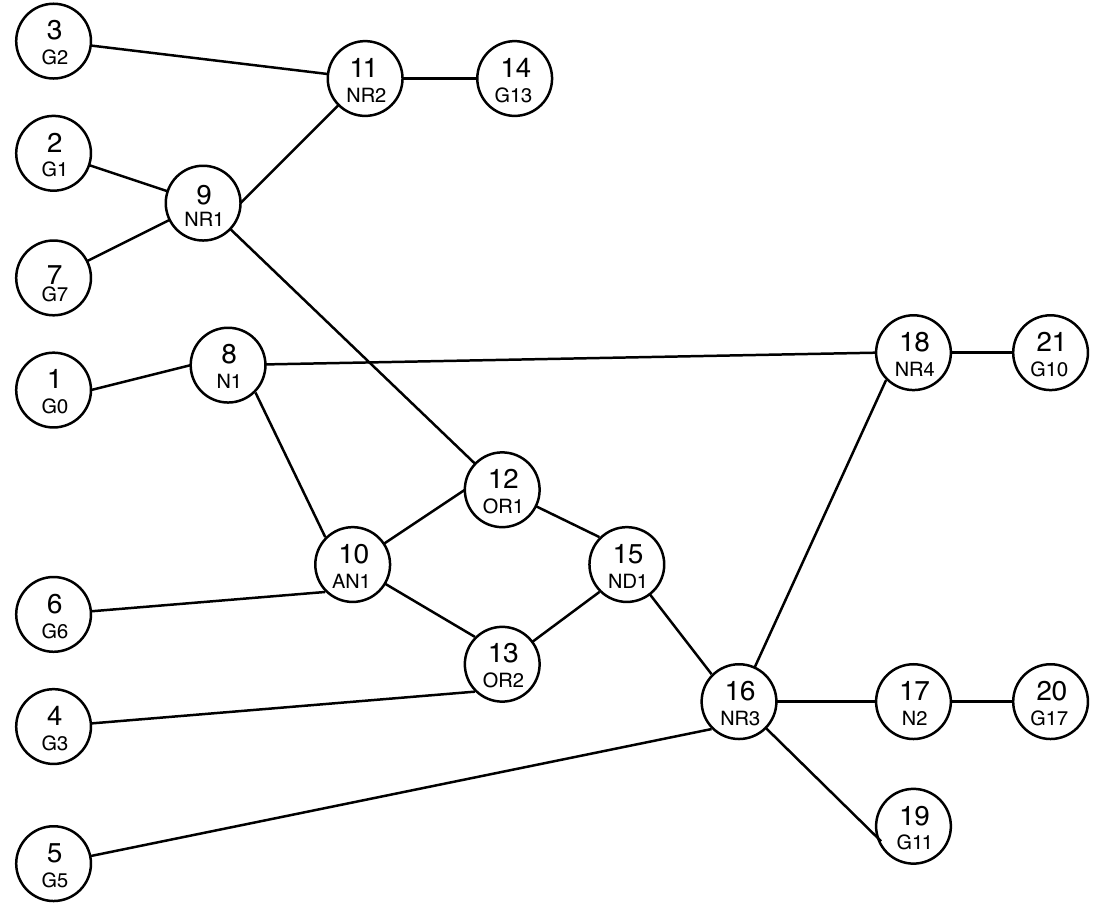}};
\end{scope}
\begin{scope}[xshift=11cm,yshift = -2.8cm]
\node {(c) circuit-graph};
\end{scope}
\begin{scope}[xshift=12cm,yshift = -4cm]
\node {\includegraphics[angle=90, height=1.5cm, width=0.5cm]{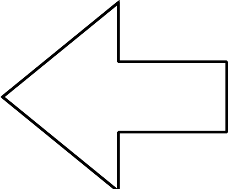}};
\end{scope}
\begin{scope}[xshift=10.3cm,yshift = -3.8cm]
\node {$\mathcal{GT}$-enhancement};
\end{scope}
\begin{scope}[xshift=1.5cm,yshift = -7cm]
\node {\includegraphics[scale =0.5]{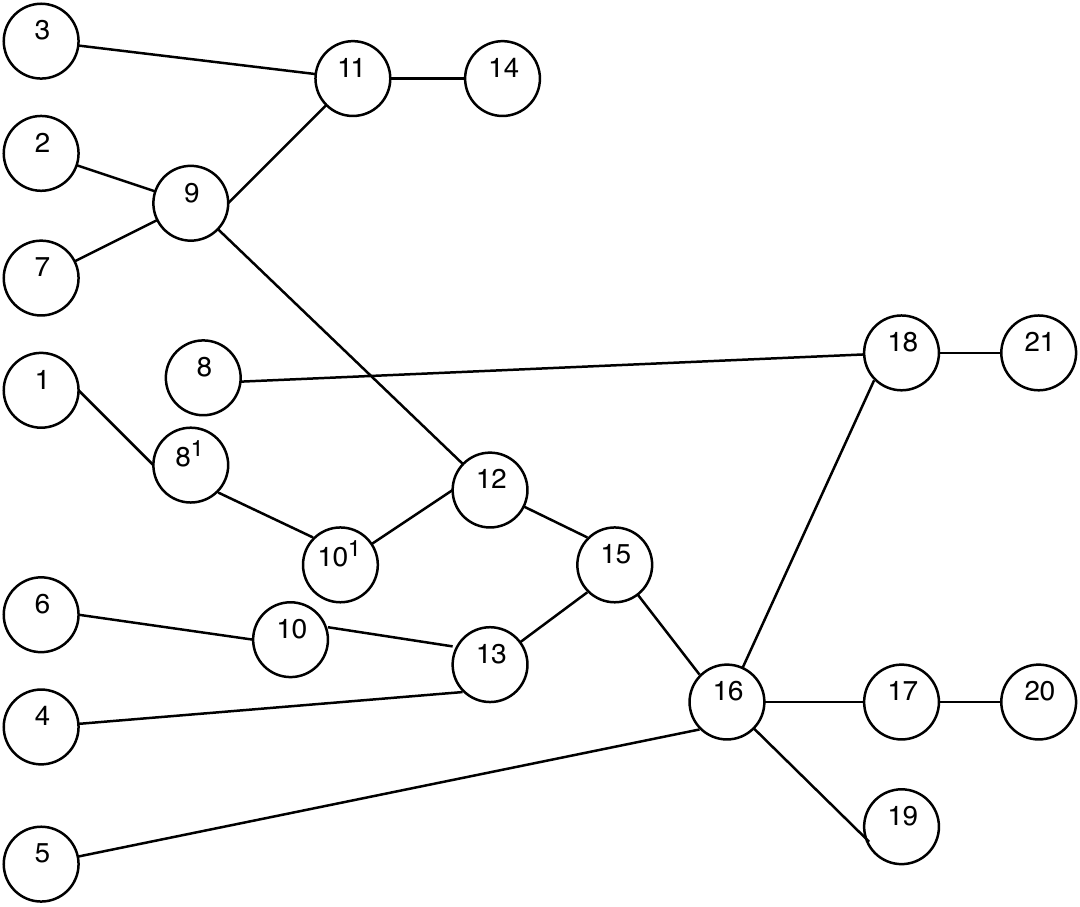}};
\end{scope}
\begin{scope}[xshift=1.5cm,yshift = -9.8cm]
\node {(f) decoded tree};
\end{scope}
\begin{scope}[xshift=11cm,yshift = -7cm]
\node {\includegraphics[scale=0.5]{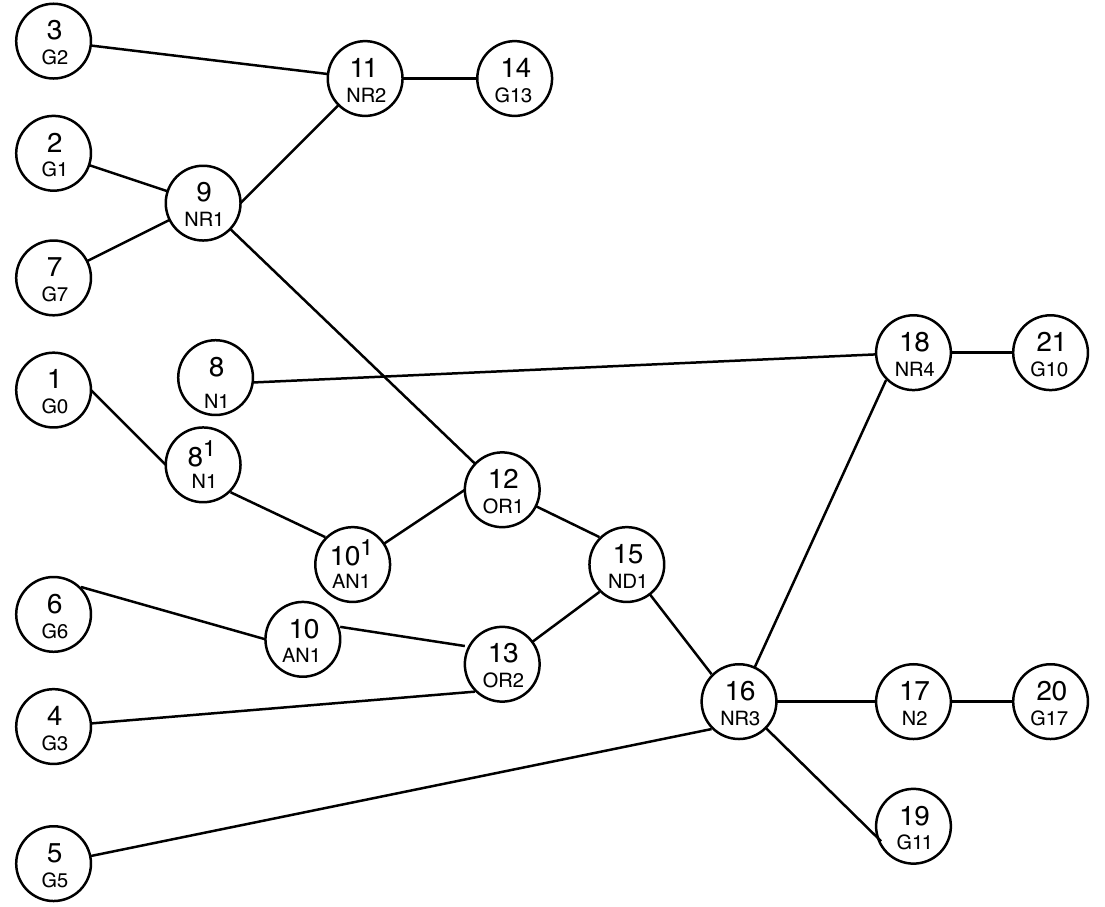}};
\end{scope}
\begin{scope}[xshift=11cm,yshift = -9.8cm]
\node {(d) \emph{g-tree}};
\end{scope}

\begin{scope}[xshift=11cm,yshift = -11cm]
\node {\includegraphics[scale =0.25]{figures/s27/arrow_turn.pdf}};
\end{scope}
\begin{scope}[xshift=6.5cm,yshift = -11cm]
\node {\includegraphics[scale =0.35]{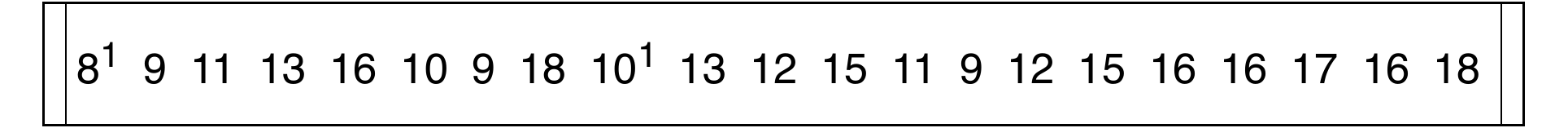}};
\end{scope}
\begin{scope}[xshift=6.5cm,yshift = -11.6cm]
\node {(e) Pr{\"u}fer code};
\end{scope}
\begin{scope}[xshift=2cm,yshift = -11cm]
\node {\includegraphics[angle=270,scale =0.25]{figures/s27/arrow_turn.pdf}};
\end{scope}
\end{tikzpicture}
\caption{An illustration of  Pr{\"u}fer code for the circuit-graph representing the benchmark circuit s27, and reconstructing it from the Pr{\"u}fer code.}\label{fig:s27_flow}
\end{figure*}

Although prufer encoding elegantly solves a major issue, the problem is yet not entirely solved because Prufer apply only to trees, that i.e., only  connected acyclic undirected graphs. 

A simple extension of  Pr{\"u}fer codes to graphs would be to partition it into a set of trees so that the graph can be represented  as the corresponding set of Pr{\"u}fer codes for individual trees. This problem does not,however, serve the purpose. The problem of partitioning the edge set $E$ of a graph $G$ into minimum number ($k$) of  trees is known to be  NP-hard  \cite{BBGD07}. Moreover for sparse graphs, even  $k$ can be very large (Figure \ref{fig:tree_partition}(a)). Also if the (sub) graphs are dense, the number of trees could be large (Figure. \ref{fig:tree_partition}(b)). For a complete graph $K_n$, the minimum number of tree-partitions is $\ceil{n/2}$. So,  the number of prufer codes required to represent a graph  being close to  $|V|$,  would be of no value. So, we need a method which completely obviates  the need of such huge number of codes. We do so by enhancing the graph into a tree by addition more vertices. This method is called  $\mathcal{GT}$-enhancement. The tree is called a \emph{g-tree}. Also, the \emph{$\mathcal{GT}$-enhancement} of $G$ preserves the number of edges of $G$ in \emph{g-tree}. Thus, the size of the Pr{\"u}fer code of a \emph{g-tree} is $|E|-1$. 

An illustration of the preservation of structural property of a graph is shown in Figure \ref{fig:s27_flow}. Here, we have taken a circuit-graph as an example and have shown its prufer code and its reconstruction. We have considered  an ISCAS'89 \cite{iscas89} benchmark circuit s27 for this purpose. The netlist and the corresponding circuit diagram is shown in Figure \ref{fig:s27_flow}(a) and Figure \ref{fig:s27_flow}(b), respectively. The circuit-graph generated from the netlist  is given in Figure \ref{fig:s27_flow}(c). Note that the graph has two cycles. $\mathcal{GT}$-enhancement, which adds two vertices $8^1$ and $10^1$, creates the \emph{g-tree} shown in Figure \ref{fig:s27_flow}(d). The tree is then encoded by a Pr{\"u}fer sequence as shown in Figure \ref{fig:s27_flow}(e). The tree structure can be completely reconstructed as shown in Figure \ref{fig:s27_flow}(f).

In summary, we  propose a new encoding scheme for digital networks based on Pr{\"u}fer sequence which has the following useful properties:
\begin{enumerate}[]
\item it is lossless, i.e., captures the structure of the entire graph;
\item it is memory efficient; the size of  encoding is  $\mathcal{O}(|V|)$;
\item the representation can be expressed as a single string of vertex labels;
\item it provides an efficient 1-D representation suitable for  machine-learning tools;
\item the encoding can be computed in time linear in the size of the graph.
\item it provides scope  improve interpretability and also to preserve additional properties of the graph like the directions on edges for DAGs.
\end{enumerate}

\section{Methodology}
\label{sec:methodology_ch5}
In this section we will briefly look at the Pr{\"u}fer encoding of a tree. We will also introduce the \emph{g-tree} and show how Pr{\"u}fer encoding can be applied to it.
\subsection{Pr{\"u}fer-Code}
Consider a tree, $T$, with $n$  vertices. The simplicity of Pr{\"u}fer encoding method through the tree can be easily recovered is primarily because it set the following  constraints regarding the labels attached to the nodes:

\begin{Constraint}
\label{assume:label} 
The vertices of $T$ are labelled sequentially as \{`$1$', `$2$', $\cdots$ `$n$'\}. 
\end{Constraint}
\begin{Constraint}
\label{assume:label_compare}
The vertices of $T$ are comparable such that \{`$1$' $<$ `$2$'  $<$ $\cdots$  $<$ `$n$'\}. 
\end{Constraint}
 
The steps to encode $T$ is given in Procedure \ref{pseudocode:encode}.
\begin{algorithm}[H]
\floatname{algorithm}{Procedure}
\small
\begin{algorithmic}[1]
\caption{Encoding: tree-to-code}\label{pseudocode:encode}
\State Fetch a single degree vertex, $v$, with the smallest label. Delete $v$. 
\State Write the label of the neighbour of $v$ to the right of the code.
\State Repeat Step 2 until only two vertices remain in the tree.
\end{algorithmic}
\end{algorithm}

Consider a Pr{\"u}fer code of $T$: $c_1,c_2,\cdots,c_{m}$; we make the following observation:
\begin{observation}
\label{ob:nm}
$n = m + 2$.
\end{observation}

\begin{observation}
\label{ob:vertex_in_code}
The degree of a vertex is one more than the number of times the label of a vertex appears in the code.
\end{observation}

\begin{corollary}
\label{cor:single_deg}
Labels of single-degree vertices do not appear in the code.
\end{corollary}

The steps for  decoding  Pr{\"u}fer code are given in  Procedure \ref{pseudocode:decode}.

\begin{algorithm}[H]
\floatname{algorithm}{Procedure}
\small
\begin{algorithmic}[1]
\caption{Decoding: code to tree \cite{JJGECCO01}}\label{pseudocode:decode}
\State Initialize a variable $k$ to 1.
\State Compute: $n \leftarrow m+2$; labels $\leftarrow$ \{$1 \cdots n$\}. \Comment{Observation \ref{ob:nm}.} 
\State Compute the degree of each node. \Comment{Observation \ref{ob:vertex_in_code}.}  
\State Fetch a single-degree vertex, $v$, with the smallest label. Set ($v,c_k$) as an edge of the tree. 
\State Decrement the degree of $v$ and $c_k$; increment $k$.
\State Repeat Step 4 and Step 5 until all  vertices have degree 0, except a pair with degree 1. \Comment{These form the last edge of the tree.}
\end{algorithmic}
\end{algorithm}

A linear-time encoding and a decoding algorithm for Pr{\"u}fer sequence are given in \cite{WWJSEA09}. 
Next we will see in what way the two  constraints defined on the labels aid in the decoding process. Firstly we decode the number of vertices, $n$, in the tree through the formula given in Observation \ref{ob:vertex_in_code} in step 2, of Procedure \ref{pseudocode:decode}). Due to the constraint \ref{assume:label} we are able to decode the labels of the vertices . The second constraint enforces a particular sequence of the labels, which is defined by the integer value of the labels. So, this constraints makes it possible to iteratively decode tree. 

It is important to note here, that although we could retrieve the information of the label of the vertices though Constaint \ref{assume:label}, some information about the labels can also be obtained from the code it self. Precisely stating, the labels of those vertices which are non-pendant are present in the code from Observation \ref{ob:vertex_in_code}. So, if need we can relax the first constraint. In such case we need to keep the label information of pendant vertices Corollary \ref{cor:single_deg}) in a separate list. Since this list will be required to decode the tree from the code. This is  important because in case of \emph{g-trees}, which we discuss next and is the prime aspect of this paper, we are required to relax  the first constraint.

\subsection{$\mathcal{GT}$-Enhancement and Encoding of  \textit{g-tree}}
\label{ssec:encode_gtree}
As discussed earlier we see that the best way to extend prufer encoding scheme to graphs would be to alter the graph  such that all the connectivity information of the graph can be embedded in  a single tree. So, in this section we discuss the proposed technique of altering a given graph G(V,E) into a single tree so that the prufer encoding of the graph is possible. We call this method  $\mathcal{GT}$-enhancement and the tree so obtained is called  a \emph{g-tree}.  The complete connectivity information about the graph is preserved in the \emph{g-tree} by preserving the edges of the graph.  We know that a tree is an acyclic graph. So, we convert the graph in a tree simply by removing all the cycles in the graph but without removing the edges.  This is done through the proposed technique of \emph{vertex-split} which is the central operation in   $\mathcal{GT}$-enhancement.  As the name suggest,  the vertex which is part of a cycle is split into two, so  we make the graph acyclic by actually breaking the cycle at a vertex. Next we will elaborate on  vertex-split operation.

A vertex-split operation on a  vertex $v$  splits or replicates the vertex such that a  copy of the vertex $v$ (call it $v^1$,) is added to the graph. We call this new vertex $v^1$ as a \emph{split-vertex}. Among the edges incident on $v$,  some of the  edges  are altered by making them  incident on its split-vertex $v^1$, such that the cycle formed because of any of these edges is no longer present. In this paper we propose two methods of $\mathcal{GT}$-enhancement. The methods differ  in the way the graph is traversed and also the way vertex-split operation is performed. The vertex-split considers different set of edges to the altered as will be discussed later.

 The  $\mathcal{GT}$-enhancement of $G$ do not add any new connection/edge to the graph, so the \emph{g-tree} obtain has the same number of edges as the original graph  $G$ that is $|E|$. So, the number of vertices in the \emph{g-tree} is $|E| + 1$.  Now, since $G$ originally had $|V|$ number of vertices, the number of additional vertices in the \emph{g-tree} is $|E|+1-|V|$.

We have seen how the Constraint 
\ref{assume:label} is taken care of by considering a separate list of labels of pendant vertices. We call this list L. Next we will discuss the labelling scheme for the vertices  so that the     Constraint 
\ref{assume:label_compare} made by the prufer encoding scheme is taken care of.  Let the \emph{g-tree} be denoted as a tree $T_g$ with vertex 
set  $V^T$ and edge set  $|E^T|$. Let the set of new vertices be  $V'$. 

 So, $V^T = V \cup V'$ and   is equal to $|E|$.  Let  $n$ denote the number of  vertices in  $T_g$,  $n_1$ denote the number of vertices in 
 the original graph and let $n_2$ denote the number of new vertices added. So, $n = n_1 + n_2$ and  $n_2 = |E| + 1 - |V|$.  So, for 
 labelling the tree $T_g$, firstly label of the vertices in $V$ are preserved as \{`$1$' $\cdots$ `$n_1$'\}.  Now, a vertex in the graph can 
 ve vertex-split to more than one vertices  during $\mathcal{GT}$-enhancement, so when any vertex  $v$ with label $l$, where $l$ in an 
 integer, is replicated into $k$ additional vertices, we label   as $l^1,l^2, \cdots, l^k$. In order to preserve Constaint 
 \ref{assume:label_compare}, which enforces some ordering among the vertex labels, while encoding $T_g$ with a Pr{\"u}fer code,  the labels 
 of these vertices are considered in the order  $l<l^1<l^2< \cdots<l^k<$$L+1$ .  

Next we discuss the process of decoding the labels from a given prufer encoding of $T_g$. The Constraint \ref{assume:label}  partially holds 
for $T_g$, since it is true only for $n_1$ vertices. So, if the information about $n_1$ is known, the labels of these  vertices can be 
computed. However, since we have added new verices here, $n_1$ cannot be directly computed.  The only information we can come is the value 
of  $n$ (Observation \ref{ob:nm}). Next, we need to compute $n_2$ from which we get the value of $n_1$ .  Now,  to compute $n_2$ we need to 
know the information about the labels of the vertices in $V'$. However, for Corollary \ref{cor:single_deg}, not all labels appear in the prufer cod. So as discussed earlier from the list $L$ and prufer code be obtain the 
labels of vertices in $V'$ and also its size $n_2$.

In the next section,  we discussed the  two  approaches to  $\mathcal{GT}$-enhancement. As we earlier, a drawback of this technique is 
the we need to store an additional list. So, The first method aims at reducing the list $L$ such that $|L| \ll |E|$.  The second method 
completely obviates the need to store the additional list. The second method produces the \emph{g-tree} where all the additional vertices are 
of degree two and the labels of vertices in $V'$ appear in the code; thus the Pr{\"u}fer code thus obtained completely represents the graph.

\section{Tree-Partition Based $\mathcal{GT}$-Enhancement}
\label{sec:gt-tree_based}

In this method, representing an undirected graph by a tree is realized by partitioning the edges of the graph into trees. Thereafter, the individual trees are joined to from a single tree.

\subsection{Proposed Approach}
Since the problem of finding a minimum tree-partition of a graph $G$ is hard \cite{BBGD07}, we follow a  greedy approach based on  depth-first search (DFS) \cite{Cormen}  traversal  of $G$. DFS of an undirected graph produces a  spanning tree $T_{DFS}$; the edges of this tree called are tree-edges and they form the primary partition. The rest of the edges of the subgraph $G^c = G \setminus T_{DFS}$ are called back edges. While the DFS spanning tree is implicitly constructed during the  traversal, the residual graph $G^c$ is often found to be disconnected. The secondary partitions, which are formed by the back edges of $G^c$, are called \emph{be-trees} and belong to one of the following: 
\begin{enumerate}
\item Class-1 back-edge tree (\emph{be-tree-1}): These are trees formed by the back edges between the vertices, which have already been visited once.

\item Class-2 back-edge tree (\emph{be-tree-2}): These are trees formed by the back edges between the vertices, which have already been visited twice: by  the DFS-tree and also by a  be-tree-1. 

\item $q^{th}$ class back-edge tree (\emph{be-tree-q}): The trees formed by the back edges between the vertices that have already been revisited $q-1$ times, are called  \emph{be-tree-q}. 
\end{enumerate}

 Note that the DFS-traversal usually  yields a number trees for each class. For a given class $i$,  all trees in \emph{be-tree-i}  are independent to each other since they have no vertex in common.   The \emph{be-trees} that are non-independent to a given $k^{th}$ {be-tree-1},  are said belong to the same \emph{family}, $f_k$. Thus, the edges of the graph are partitioned into a spanning DFS-tree and a set of  families of non-independent trees. 

\begin{figure}[h]
\centering
\includegraphics[scale=0.8]{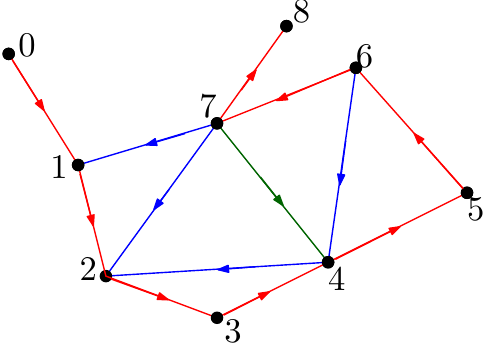}
\caption{Exmple graph showing the DFS-trees and the secondary partitions.}\label{fig:eg_graph}
\end{figure}

\emph{Example}: Figure \ref{fig:eg_graph}(a) shows an example graph, where the DFS-tree is shown in red. The decomposition consists of one family of \emph{be-trees} with two classes: \emph{be-tree-1} (blue) and \emph{be-tree-2} (green). Also note that in Figure \ref{fig:tree_partition}(a), the graph has eight families of be-trees.

Next we look at the way the process of $\mathcal{GT}$-enhancement is applied in this method. 
 Given a graph $G(V,E)$ and the set $P$ of secondary partitions, for each partition in $P$, a \emph{join vertex} ($v_{join}$) is specified. $G$ is decomposed into a \emph{g-tree}, $T_g$, based on $P$ and their  join vertices. Each vertex in $V$, belongs to the primary partition. For every vertex $v$ that belongs to some secondary partition $P_j$, a \emph{vertex-split} operation (refer to Figure \ref{fig:eg_split}(a)) is performed. The vertex $v$  is split into a split-vertex  $v^{P_j}$ for every partition $P_j$ where it belongs to, such that the edges of $G$ that belong to $P_j$ and were incident on $v$, become incident on $v^{P_j}$. However, if this vertex is a join-vertex of a partition,  its replica  for that partition is not created. Thus, the tree $T_g$ is formed by the trees for each partition in $P$, each tree sharing a common vertex with the spanning tree at its join vertex. We called this tree \emph{DFS-Partition-tree} and denote it as $T^{DP}_G$. The labeling methodology for the new vertices was described previously in Section \ref{sec:methodology_ch5}.

\begin{figure}[h]
\centering
\includegraphics[scale=0.8]{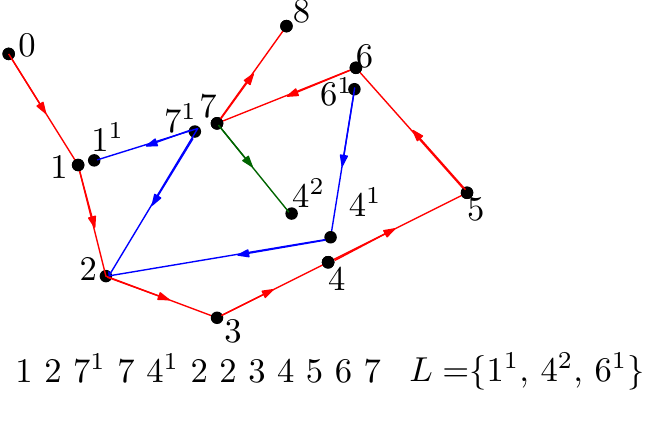}
\caption{$\mathcal{GT}$-enhancement: \emph{g-tree} of the example graph. The prufer code and the list L is also shown in the figure.}\label{fig:eg_before_split}
\end{figure}

\emph{Example}: On performing $\mathcal{GT}$-enhancement on the graph in Figure \ref{fig:eg_graph}, we obtain the  resulting graph as shown in Figure \ref{fig:eg_before_split}. 

As shown in Figure \ref{fig:eg_before_split}, there are there pendant split-vertices in the \emph{g-tree}. So, we need to store their label in the list L. So, next we will look at some operations that can be incorporated in the traversal such that the size of L can be reduced.

\subsection{Reducing the size of the list L}
Firstly we will see a method to reduce the number of \emph{be-tree} classes. Since the number of split-vertices is governed to the number of  \emph{be-tree} classes that it belongs to, so this process is favorable at reducing the number of labels that needs to be possibly stored. We call this method as \emph{edge-swap} and is explained as follows.
Consider a pair of vertices, $u$ and $v$, of $G$ such that they are connected by an edge $e$ and also by two paths $p_{dfs}$ and $p_{be1}$. Suppose  DFS exploration, traverses  the path $p_{dfs}$, including it in the DFS-tree, such that $v$ is visited  before $u$.  Let $p_{be1}$  be assigned to  \emph{be-tree-1}, $b_1$. We  define two special edges:\\
\emph{Cycle-edge:} At some instant during the traversal when vertex $u$ is being processed, let edge $e$   be a back edge from $u$ to $v$. Since it will  introduce a cycle in  $b_1$, it is assigned to  \emph{be-tree-2}. Such an edge is called a cycle-edge.\\
\emph{Swap-edge:} Now, suppose there exists a tree-edge $e_s$, in the path $p_{dfs}$, which  does not form a cycle with any \emph{be-tree-1}. Then the edge $e_s$ is swapped with $e$; $e_s$ is then  called a swap-edge for the cycle-edge $e$.\\
\emph{Edge-swap:} An edge-swap between the cycle-edge $e$ and its swap-edge $e_s$ enables $e$  being assigned to a tree-edge and $e_s$  being assigned to  \emph{be-tree-1}.  A new class of \emph{be-tree} for the cycle-edge $e$ is thus  avoided by an edge-swap operation. Note that an edge-swap, modifies the DFS-tree, without disconnecting it.

\begin{figure}[h]
\centering
\includegraphics[scale=0.8]{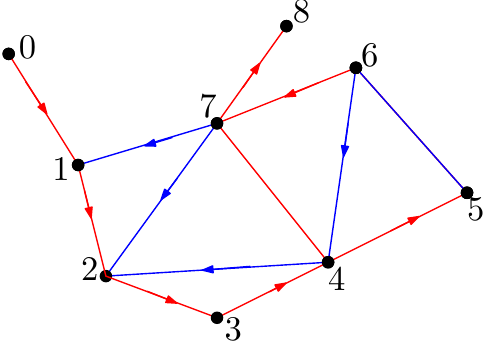}
\caption{Edge-swap}
\label{fig:edge_swap}
\end{figure}

\emph{Example}: Consider the graph in Figure \ref{fig:eg_graph};  edge (7-4) and edge(5-6) would be a cycle-edge and swap-edge, respectively. The resulting graph is shown in Figure \ref{fig:edge_swap}.

There are scope to further reducing the list $L$ by taking care of how we form \emph{g-tree} so that we can smartly handling the labels.
Because reducing $L$ imply ensuring that the new labels to point to non-pendant vertices.  So, the following two operation exploit this to reduce the size of L:
\begin{enumerate}
\item \emph{Choice of join-vertex}: It is so chosen  that it becomes a pendant vertex of the \emph{be-tree} and  the DFS-tree (if possible);
\item \emph{Label-swap}: The replicas of a vertex in $v$ have similar labels; so exchanging the labels does not change the structure of $G$. The operation label-swap is applicable to a pair of vertices $(v,v^r)$ with corresponding label $(l,l^r)$, where $v^r$ is a replica vertex of $v$, $v$ is a non-pendant vertex and $v^r$ is a pendant vertex. In such case their labels are swapped. So, the label $l^r$  now points to a non-pendant vertex and hence it does not need to be stored.  So, this operation decreases the size of $L$.

\end{enumerate}

\begin{figure}[h]
\begin{subfigure}{0.5\textwidth}
\centering
\includegraphics[scale=0.6]{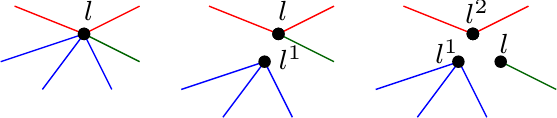}
\caption{Vertex split with label-swap}
\end{subfigure}

\begin{subfigure}{0.5\textwidth}
\centering
\includegraphics[scale=0.8]{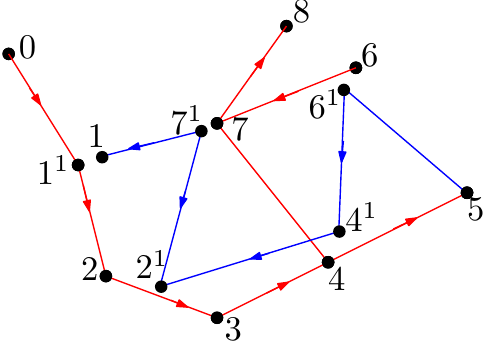}
\caption{\emph{g-tree} for the graph in Figure \ref{fig:eg_graph} \& its Pr{\"u}fer sequence}
\end{subfigure}
\caption{\emph{g-tree} and label-swap operation}\label{fig:eg_split}
\end{figure}

Example: The  Figure \ref{fig:eg_split}(a) depicts label-swap. Also  in Figure \ref{fig:eg_split}(b), label-swap is performed for the vertex with label ``1". The  vertex with label ``5" is chosen as join-vertex since it is  pendant in both the trees it belongs to. Hence, after all  operations, the list $L$  reduces to  $\phi$. The corresponding Pr{\"u}fer sequence is shown in the figure.

\subsection{Implementation}
In this section, we discuss the outline of our implementation on benchmark integrated circuits \cite{iscas89,itc99}. Since scan-based circuits are envisaged as directed acyclic graphs, we first label their nodes
in topologically-sorted order \cite{KACM62} and run a single DFS-traversal, considering it as an undirected graph (directions can be retrieved from the labels). Also, the successors of each vertex are visited in the increasing order of their labels. When an edge $e(u,v)$ is discovered, it is (i) assigned to a particular partition by an operation called add-edge (described later), or (ii) added to the list of cycle-edges, or (iii) assigned to a partition (add-edge) if it is already marked as a cycle-edge and has not yet been swapped. When an edge is backtracked, we check whether it is suitable for swapping with a swap-edge in the list. Lastly, once all the edges emanating from a vertex are processed, vertex-splitting along with label-swap is performed on it if needed.

During each successive iteration of the DFS-traversal, search is made to explore an unvisited vertex, and if found, the connecting edge is added to the DFS-tree. Similar implicit exploration cannot, however, be  performed to extract \emph{be-trees}. A pair of edges in a \emph{be-tree}, or the edges of the successive classes in a family of \emph{be-trees}, could be discovered during two independent iterations. Thus, to keep track of \emph{be-trees}, we record the following entities, which are updated whenever a new back-edge is discovered: (i) \emph{tree index}: every back-edge that does not share a common vertex with any other \emph{be-tree} formed so far, or form a cycle with the adjacent \emph{be-trees}, is considered a new \emph{be-tree} with one-higher index; (ii) vertex-attribute for each vertex $v$ (tree-index list $\mathcal{T}_v$): a ordered tuple of indices of different classes of \emph{be-trees} to which the vertex belongs, and  (iii)  edge-attribute for each edge e ($c_e$): the \emph{be-tree} class index.   $\mathcal{T}_v[c_e]$ gives the tree-index of the $c_e^{th}$ class.     

\input{tex_files/tree_part_table}

\emph{add-edge}: When a unvisited back edge  $e(u,v)$ is encountered with tree-index lists $\mathcal{T}_v$ and $\mathcal{T}_u$, $c_e$ is computed from the two lists  such that it does not form a cycle with any of the trees to which $v$ and $u$  belong. Thereafter, it is assigned to a \emph{be-tree} by updating the lists  depending on the following three cases; (i) union of two \emph{be-trees}: Suppose $e$ connects two back-edges $e_1(v,w_1)$ and $e_1(w_2,u)$, each belonging to distinct \emph{be-trees} with tree-indexes $i_1$ and $i_2$, respectively, but both having the same tree-class. In such a case, the three edges are assigned to a single \emph{be-tree} having tree-index $i_1$. This is done by treating the tree-indices as disjoint sets. Using the disjoint-set data structure, we perform union($i_1$, $i_2$); (ii) if only one of the two vertices belongs to a unique class of \emph{be-tree}, the edge $e$ is assigned to that \emph{be-tree}; (iii) lastly, if it does not fall into the above two cases, it creates a new \emph{be-tree} as explained above.

The time complexity of the algorithm is $\mathcal{O}(|V|+|E|)$.

\subsection{Results on Benchmark Circuits}
Results for   ISCAS'89 and ITC'99 benchmark-suites are shown in Table \ref{table:pruf_experimental_result}. The experiments were carried out on an Intel Xeon 3.00-GHz $\times$ 4 processor with 8GB memory.  Each of the benchmark circuits corresponds to the netlist of a digital logic circuit, where the input-ports/output-ports, logic gates, and memory cells are represented as vertices, and the interconnections among  them, as edges, of a directed acyclic graph. The circuit-name, the number of vertices, and edges are
given in Columns 1, 2 and 3, respectively. These graphs are very sparse as the edge-count is just around two times of the vertex-count. The length of Pr{\"u}fer code is given in Column 4. The required number of extra-labels $(L)$ is given in Column 5, 
which is much less than the number of edges. The highest-index of \emph{be-trees} is given in Column  6, and the number of edge-swaps is given in Column 7. The number of vertex-splits is shown in Column 8, and out of it the number of vertices whose labels are swapped, is shown in Column 9. The CPU-time needed to encode the circuit using Pr{\"u}fer-sequence is reported in Column 10, in seconds.

\section{SCESOR $\mathcal{GT}$-Enhancement}
\label{sec:gt-setsor}
\begin{figure}[h]
\begin{subfigure}{0.17\textwidth}
\centering
\includegraphics[scale=0.5]{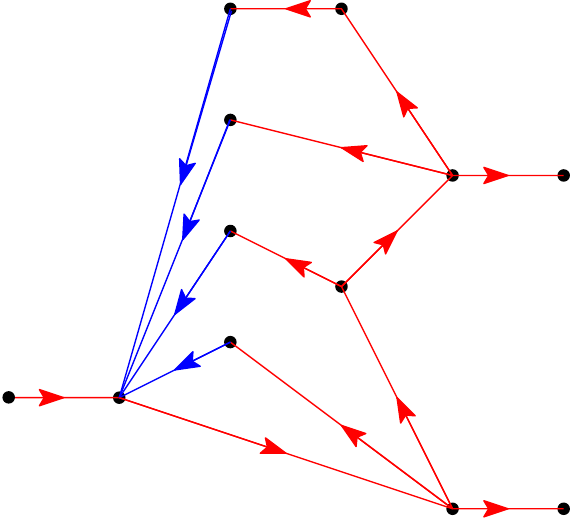}
\caption{}
\end{subfigure}\begin{subfigure}{0.17\textwidth}
\centering
\includegraphics[scale=0.5]{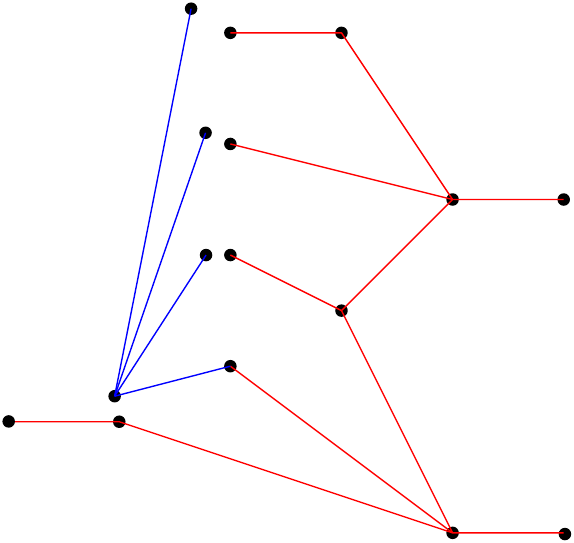}
\caption{}
\end{subfigure}\begin{subfigure}{0.16\textwidth}
\centering
\includegraphics[scale=0.5]{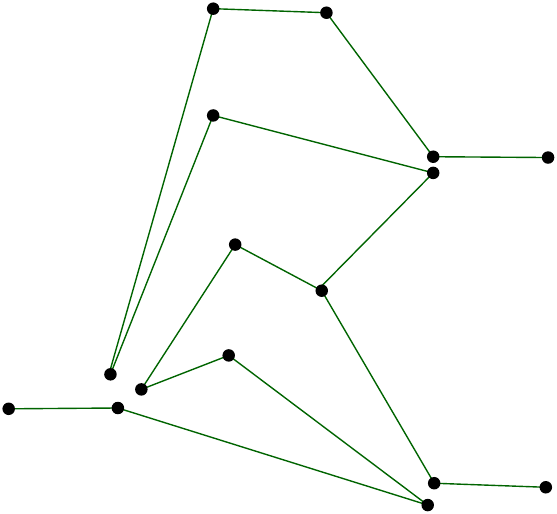}
\caption{}
\end{subfigure}
\caption{(a) Sub-graph showing special case; (b) Vertex-split by tree-based $\mathcal{GT}$-enhancement showing three new pendant vertices that can not be label-swapped; (c) Better method of $\mathcal{GT}$-enhancement with no new pendant vertices}\label{fig:tree_part_issue}
\end{figure}

The previous  method of  tree-partition based method for $\mathcal{GT}$-enhancement is designed solely too meet the purpose of converting the graph the \emph{g-tree} by breaking every  cycle by applying the method of vertex-split. It accomplishes the task through  a single DFS traversal taking  linear time. The propensity to fulfill the primary objective have made way to some pitfalls related to optimization of various factors. These are discussed as follows.
\begin{enumerate}
\item The Prufer sequence is not enough to decode the graph. As seen in Table \ref{table:experimental_result}, although the methods tries to reduce the size of the list $L$, for almost all the circuit-graphs, it could not entirely preclude its necessity. Clearly, for  decoding the graph, this methods can not solely rely on the  Prufer code. The graph traversal process was not exploited fully to meet this end since we just followed the conventional "dept-first'' approach. From ML point of view, a single string would be more convenient
for good representation of data. However, this method do not consider it as one of the  primary objectives.
\item Supplementary operations required to reduce the size of $L$. The vertex-split operation of the this method of $\mathcal{GT}$-enhancement do not  target to ensure that the new vertices added are non-pendant. It relies on a set of operations like label-swap or edge-swap. These operations lends undesired complexity to the algorithm.
\item In the cases where a vertex has high degree and that of the 
most of its neighbors are two, then the number of single-degree 
nodes which cannot be label-swapped could be large. An instance of this is shown in  the Figure \ref{fig:tree_part_issue}.  The sub-figure \ref{fig:tree_part_issue}(a)  shows a sub-graph of an instance of such case. The edges belonging to the DFS-tree is shown in red and the edges belonging to a be-tree are shown in blue. Sub-graph \ref{fig:tree_part_issue}(b) shows the \emph{g-tree} obtained by the tree-partition based method. As shown in the figure, there are three pendant split-vertices which can not be label-swapped. This mandates their appearance in the list  $L$. Thus making the need of the list $L$ unavoidable. We observe that such split-vertices can be avoided even for such tricky cases by following a different method of traversal and vertex-split through  which we arrive at a distinct \emph{g-tree} where the split-vertices and no longer pendant. This is shown in sub-figure \ref{fig:tree_part_issue}(c). This \emph{g-tree} is obtained by the second method of $\mathcal{GT}$-enhancement called SCESOR $\mathcal{GT}$-enhancement. We will now discuss this method.

\end{enumerate}

The second method of  $\mathcal{GT}$-
enhancement overcomes these shortcomings.
We call this method Seek-Clear-Edge-and-Split-On-Revisit (SCESOR) $
\mathcal{GT}$-enhancement. This method  maintains that any 
split-vertex that is added during $\mathcal{GT}$-enhancement is always of degree 
two.  Thus, this method obviates the need  for the list  $L$. Thus, 
{\it g-tree} can be represented by a single Pr{\"u}fer code. Furthermore, 
the method is much simpler to implement. 
The SCESOR  $\mathcal{GT}$-enhancement involves processes:

\begin{enumerate}
\item Seek-clear-edge graph traversal.
\item Split-on-revisit of every vertex.
\end{enumerate}

We will discuss each of these process in the following sections.
 
\subsection{Seek-Clear-Edge (SCE) Traversal}
Here we propose a novel traversal technique which is called Seek-Clear-Edge traversal or SCE traversal.
The two classical graph-traversal algorithms, which is  as DFS and BFS (breadth-first-
search),  follow an exploration method that is vertex-centric. The term ``search" 
refers to visiting the vertices sequentially; where the search techniques, which is either
``depth-first" or ``breadth-first", refer to the vertices. In contrast,  traversal scheme we propose edge-centric; 
where the exploration will be guided by edges rather than vertices. The SCE-algorithm  of a graph $G$  is  as follows. Starting from a vertex 
$v \in G$,an unvisited edge adjacent to $v$, $e^1(v,w)$ is visited (flagged). Iteratively, the  next unvisited edge adjacent to $w$ is visited. During the traversal, when no  unvisited/clear  adjacent-edge is available from the current vertex $u$,  then the algorithm  backtracks to its parent vertex $u_p$. If there is any unvisited edge adjacent to $u_p$, it is iteratively visited, and the process is continued until all edges in G are flagged.

This traversal algorithm differs with the classical approach in the following ways:
\begin{enumerate}
\item A vertex  is treated  the same; both when discovered for the first time or is when revisited. 
\item Backtrack from a vertex only when all its adjacent edges have been visited.
\item Its follows dept-first approach but goes deeper than DFS since its backtrack criteria is fully relaxed.
\item Unlike DFS, here all the edges are tree-edges and so there are no back-edges. 
\end{enumerate}

The central idea of this traversal algorithm is that at any instance during the traversal, when a clear edge is available from the current vertex, the edge is visited. Thus, the traversal is regulated by the availability of clear adjacent edges and 
not by the vertices as typical of the classical traversal process. 

\begin{figure}[bh!]
\begin{subfigure}{0.25\textwidth}
\centering
\includegraphics[scale=1]{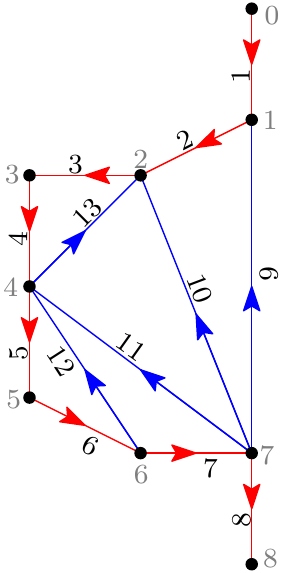}
\caption{DFS}
\end{subfigure}\begin{subfigure}{0.25\textwidth}
\centering
\includegraphics[scale=1]{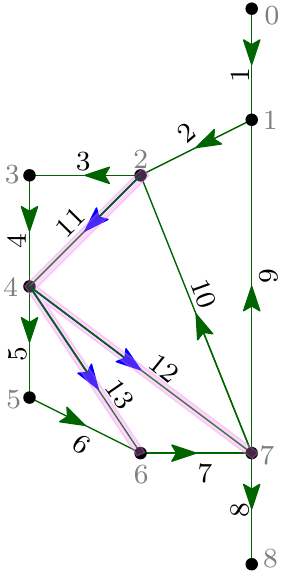}
\caption{SE}
\end{subfigure}
\caption{The graph  in Fig. \ref{fig:eg_graph} is redrawn here to show the  difference in the sequence of edge-traversal. Three  edges are shaded in 
(b) to highlight their difference. }
\label{fig:eg_graph_fet}
\end{figure}

An example of the traversal on the graph in Fig. 
\ref{fig:eg_graph} is given in Fig. \ref{fig:eg_graph_fet}. The graph is redrawn 
for clarity. The edge-labels depict the sequence number in which the edges are 
traversed. The arrow heads show the direction of traversal. Fig.
\ref{fig:eg_graph_fet}(a) shows the DFS traversal sequence and Fig.
\ref{fig:eg_graph_fet}(b) shows the SCE-sequence. Since the adjacent edges of a vertex can be chosen for traversal in any order, we assume that this order is same for both traversals for the sake of comparison. So, the edge sequence is identical up to 10. The $10^{th}$ edge is (7,2), which is incident on vertex 2. The next unvisited edge adjacent to 2 is (2,4.) So, in SCE, the $11^{th}$ edge traversed is (2,4). Note that, although the edge (2,4) is readily traversable, DFS does not allow traversal of this edge, and instead backtracks. In fact, DFS processes the edge (2,4) in the end after processing (4,6) and (4,7). Thus, SCE is more convenient for systematic exploration of edges.
 
The need for such kind of traversal technique arises because here we aim at ensuring that the split-vertex are not pendant. To this end, it would be rational that when we revisit a vertex, we check for its adjacent clean edges so that we have better choice while performing vertex split. In case of DFS, we arrowed to have access to other clean adjacent edges , and most importantly know about such edges, only  on successive revisits. Since, we know about the adjacent vertex in advance, relative to DFS, this method is also computationally better and a lot simpler in addition to it being efficient. In fact, it implicitly ensures that the split-vertex is of minimal degree, that is, of degree two. In the next section, we will discuss the vertex-split approach called the split-on-revisit (SOR).

\subsection{Split-On-Revisit (SOR)}
SOR is a simple  mechanism for vertex-split that can be easily incorporated in the SCE traversal. It is explained as follows. During the traversal, if a vertex  $v$ with label $l$ is revisited through an edge $e_{cycle}$, it implies that it is a part of a cycle. To break the cycle, $v$ is split as follows:  a new a vertex $v^1$, labeled $l^1$, is created.  During the previous visit to $v$, let $e_{in}$ be the edge through which it was visited and let $e_{out}$ be its adjacent edge which was next traversed. Obviously, $e_{out}$ is also a part of the cycle. So, the pair of edges ($e_{in}$, $e_{out}$) is now connected to $v^1$ instead of $v$. The edge $e_{in}$ maintains the connectivity to the edges traversed up to it while the edge $e_{out}$ contributes to disconnecting the cycle and the maintenance of connectivity to the rest of the edges. We call  ($e_{in}$, $e_{out}$)  the \emph{split-pair} edges of $v$.
\begin{figure}[h!]
\begin{subfigure}{0.16\textwidth}
\centering
\includegraphics[scale=0.5]{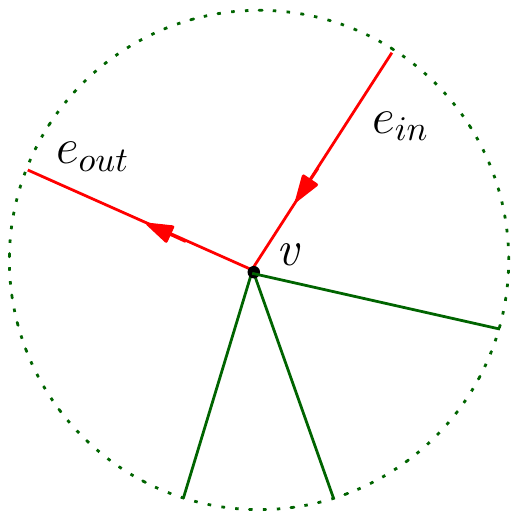}
\caption{}
\end{subfigure}\begin{subfigure}{0.16\textwidth}
\centering
\includegraphics[scale=0.5]{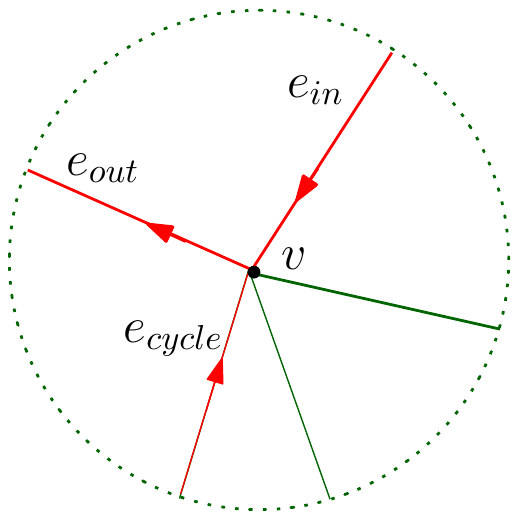}
\caption{}
\end{subfigure}\begin{subfigure}{0.16\textwidth}
\centering
\includegraphics[scale=0.5]{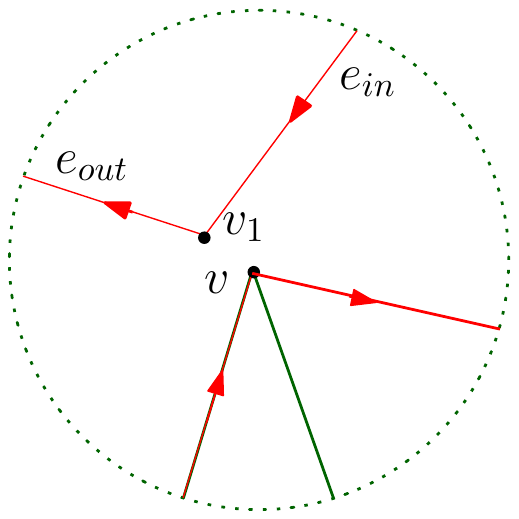}
\caption{}
\end{subfigure}
\caption{An illustration of the SOR operation. The edges colored in red depict those edges which have been traversed. It shows the three instance which contribute to the vertex-split. Figure (a) shows the split-pair edges when a vertex $v$ is visited prior to the split corresponding to this split-pair edges. Figure (b) shows the instance when the vertex is revisited through the edge $e_{cycle}$. Figure (c) shows the vertex-split leading to the introduction of the split-vertex $v_1$.}
\label{fig:eg_sor}
\end{figure}
This is illustrated in the Figure \ref{fig:eg_sor}.
On  every successive revisits to $v$, it is split similarly, and the $i^{th}$ revisit will produce a replica which is labeled as $l^{i}$. Thus, the degree of $v$ is reduced by two in every revisit. Such a traversal breaks all the cycles while maintaining the connectivity of the graph  turning it into a tree. We call this method \emph{Split-On-Revisit} or \emph{SOR} and the \textit{g-tree} obtained is called \emph{SCESOR-tree}. 

 The algorithm for the SCESOR  $\mathcal{GT}$-enhancement is summarized in the pseudocode  given in Algorithm \ref{pseudocode:split-graph}. The function vertex-split $v$ mentioned in line 2 of the pseudocode refers to the SOR vertex-split operation  explained above.

\begin{algorithm}[H]
\scriptsize 
\begin{algorithmic}[1]
\caption{SCESOR($v$)}\label{pseudocode:split-graph}
\If{$v$ is visited earlier}  \hspace{4cm} $\tikzmark{top}$ 
		\State \texttt{vertex-split} $v$ 
\EndIf                                  $\tikzmark{bottom}$
\State visit $v$
\While {edge($w,v$) in unvisited adjacent edge-list of $v$} 
\hspace{0.8cm}$\tikzmark{topf}$
	
	\State \texttt{SCESOR}($w$)             $\tikzmark{bottomf}$

\EndWhile
\end{algorithmic}
\end{algorithm}

\AddNote{top}{bottom}{top}{\small{SOR}}
\AddNote{topf}{bottomf}{topf}{\small{SCE}}

\begin{figure}[h!]

\centering
\includegraphics[scale=1]{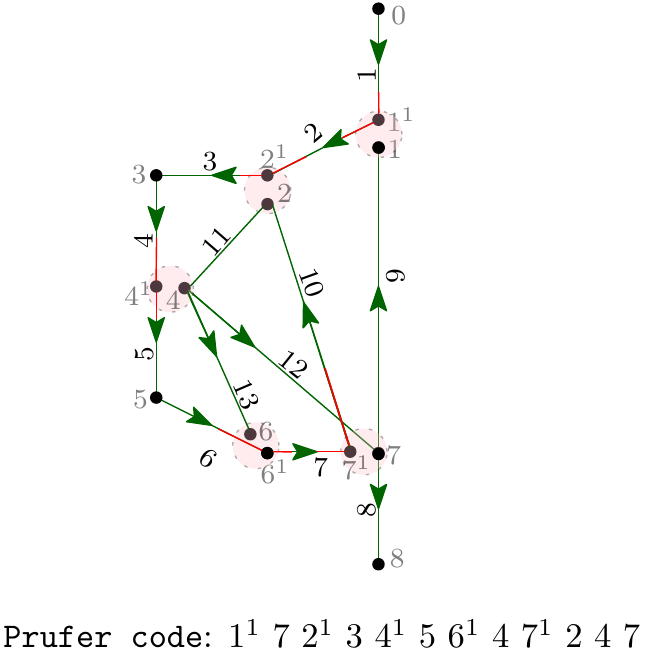}

\caption{SCESOR-tree of the graph  in Fig. \ref{fig:eg_graph_sor}(b) along with the prufer code. }\label{fig:eg_graph_sor}
\end{figure}

The SCESOR-tree of the graph in Fig. \ref{fig:eg_graph_fet}(b) is shown in Fig. \ref{fig:eg_graph_sor}. The five instances of 
vertex-splits are circled in pink shade in the figure. For vertex 1, it is 
split into vertex $1^1$ with the split-pair being ((0,1),(1,2)). Similarly, we can observe other split-pairs which are marked red in the figure. The prufer code for the \emph{gtree} is also given in the bottom of the figure. The the label of all the split-vertices appear in the code. Note that they appear exactly once in the code and this is true for an encoding of a SCESOR-tree.

A SCESOR transformation of a  graph $G(E,V)$ splits a subset of vertices in $V$. In the  SCESOR-tree thus obtained, the vertices fall into three categories, where a vertex  is given a denotation reflecting its  category. A vertex $\in V$ that is split, leading to its degree being decremented, is denoted as \emph{s}-vertex. A vertex belonging to the set of  remaining vertices $\in V$, which are not split, is denoted as \emph{g}-vertex. A replica vertex created by vertex-split is denoted as \emph{r}-vertex. We make the following observations about a SCESOR-tree. (i) An \emph{r}-vertex  is of degree two, (ii) An \emph{s}-vertex
may be of degree one or more. (iii)
If an \emph{s}-vertex is pendant then its degree in $G$ is odd.

So,  only the \emph{r}-vertices carry the new labels. From the first observation, we conclude that the new labels always appear in the Pr{\"u}fer code of the SCESOR-tree for any given graph. Hence,  the Pr{\"u}fer code alone is enough to infer the labels of the vertices following the method described in Section \ref{ssec:encode_gtree}, thus decoding the tree. 
\begin{figure*}[t]
\begin{subfigure}{0.5\textwidth}
\centering
\includegraphics[scale=0.5]{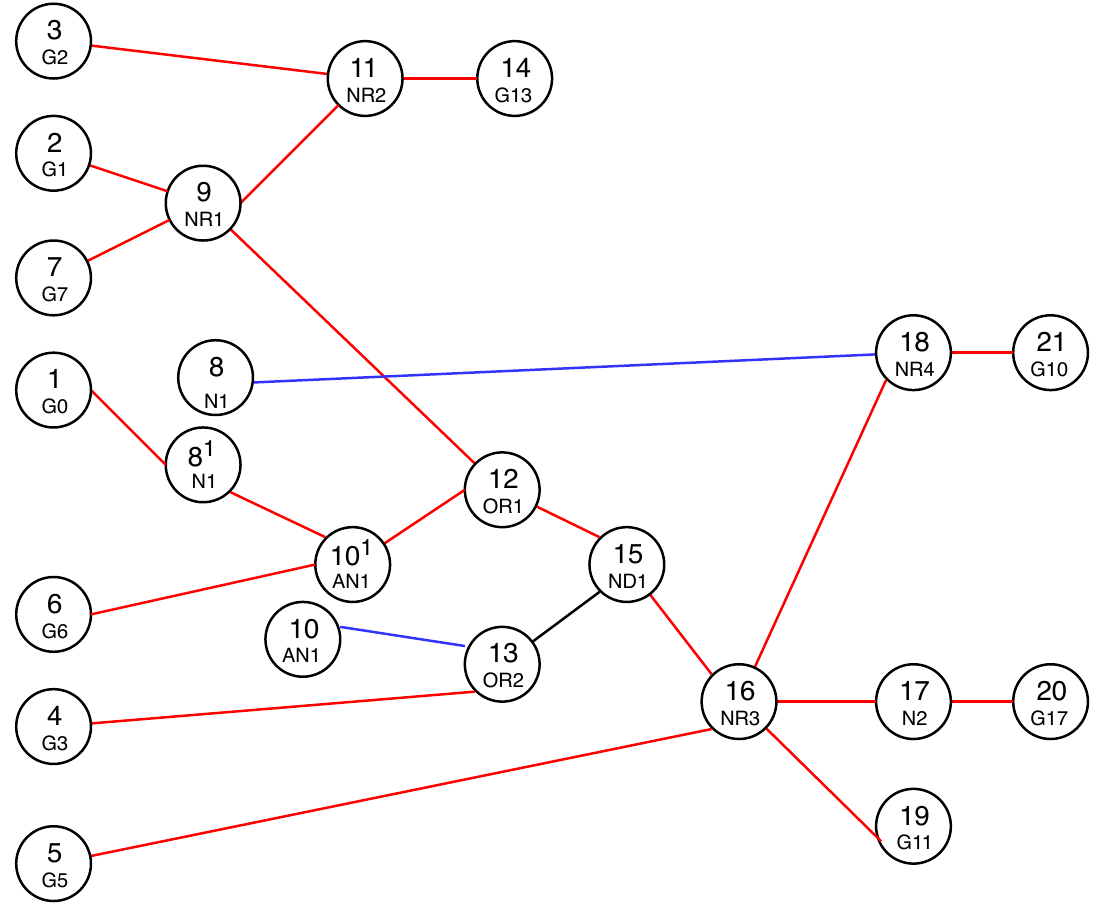}\\
{\tt Pr{\"u}fer code}:\\

\scriptsize{$8^1$   9  11  13  16  10  9  18 $10^1$  13  12  15  11  9  12  15  16  16  17  16  18 }
\caption{Tree-partition based method}\label{fig:s27_gtree_partition}
\end{subfigure}\begin{subfigure}{0.5\textwidth}
\centering
\includegraphics[scale=0.5]{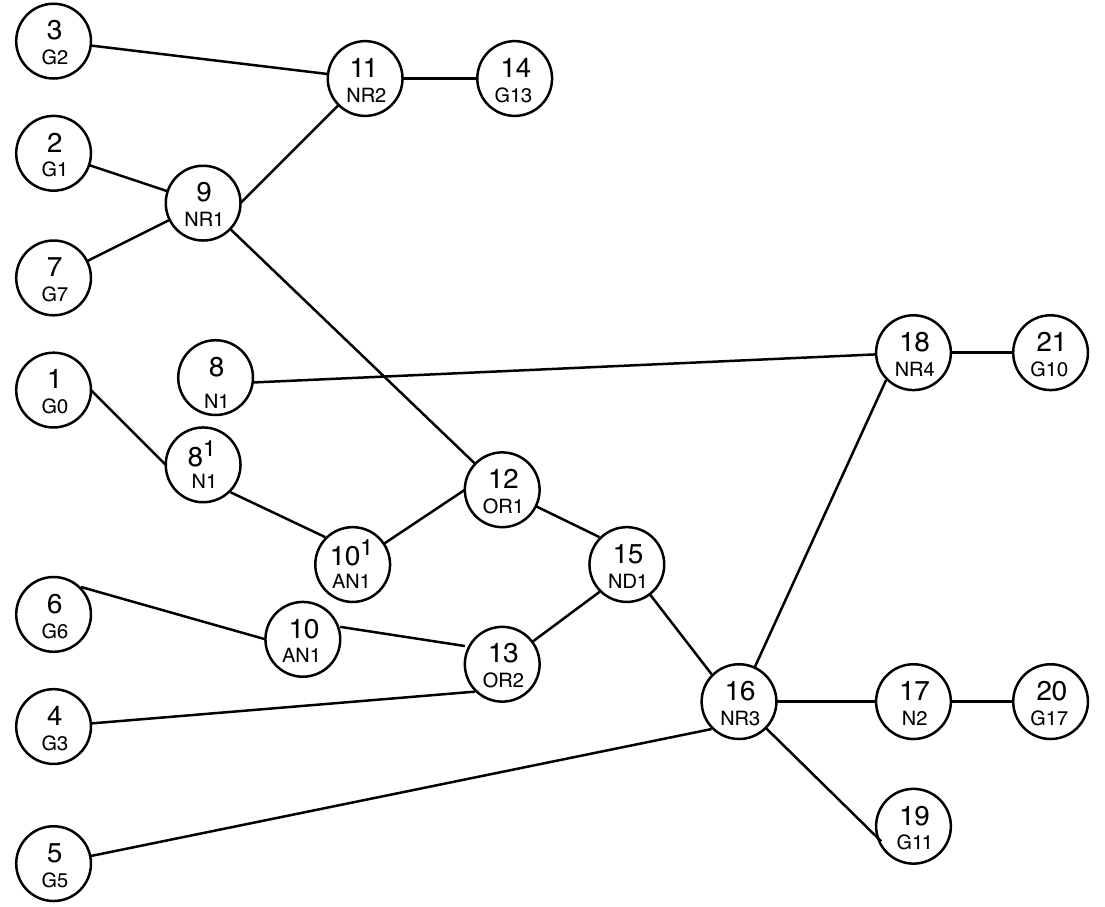}\\
{\tt Pr{\"u}fer code}:\\

\scriptsize{$8^1$   9  11  13  16  $10^1$  9  18 $10^1$  13  12  15  11  9  12  15  16  16  17  16  18 }
\caption{SCESOR-based method}\label{fig:s27_gtree_sensor}
\end{subfigure}
\caption{An example of \emph{g-tree} and Pr{\"u}fer code of s27 using both the methods}
\label{fig:s27_gtree}
\end{figure*}

The \emph{g-tree}s and the corresponding Pr{\"u}fer sequences for s27 obtained using the two approaches is given in Figure \ref{fig:s27_gtree}. The two  new replica vertices added are $8^1$ and $10^1$. Figure  \ref{fig:s27_gtree_partition} shows the \emph{g-tree} for the tree-partition-based method. There are two be-trees, marked in blue. The new vertices are pendant and so the vertex-label is swapped for both of them, making the list of extra vertices $L$ empty. However this is not the case with most of  the benchmark circuits as depicted in Table \ref{table:pruf_experimental_result}. The SCESOR-tree for s27 is given in Figure \ref{fig:s27_gtree_sensor}. It can be seen that both the  r-vertices are of degree two. As discussed, this method implicitly produces r-vertices of degree two for any given circuit-graph.

In the next section, we will discuss some methods that  improve the interpretability of a code by exploring some 
attributes of a Pr{\"u}fer code.

\section{Pr{\"u}fer Code Selection}
\label{sec:code_selection}
We have seen that the structure of a tree can be fully preserved by its Pr{\"u}fer code. Besides this primary utility of Prufer codes, another  peculiarity  about the Pr{\"u}fer encoding is that the Prufer code for a tree is not unique; a tree can be encoded into a set of distinct Pr{\"u}fer codes.  So this property offers the opportunity to make a choice of a suitable code to represent a \emph{g-tree} such that the chosen code have the desired requirements. (i) It can be chosen such that it exhibits good interpretability which is an important condition for it be be applicable for ML prupose. (ii) It can also be used so as to  preserve the attributes of the graph such as the directions of edges, and (iii) Lastly, it could offer more flexibility such it facilitates learnability of the code. Here, we present three different Prufer codes for \emph{g-tree}. Before that, we discuss some  properties of Prufer encodes to understand how it lends such variety of codes for a tree. Lastly, we discuss about the learnability of such codes.

\subsection{Properties of Pr{\"u}fer Code}
\label{subsec:pruf_prop}

Consider an unlabelled tree $T$ with $n$ vertices. 
A Pr{\"u}fer code of $T$ has the following properties:

\begin{enumerate}[leftmargin=*]
\item \textit{A Pr{\"u}fer encoding induces an edge sequence.} 

\label{pruf_prop_1}
A Pr{\"u}fer encoder monitors the list of pendant vertices. Let us call this list \emph{Pen\_List}. In each iteration, \emph{Pen\_List} is updated. From Procedure \ref{pseudocode:encode}, we recall that the basic operation in each iteration of encoding (decoding) $T$ into a Pr{\"u}fer sequence is: ``choose the pendant vertex ($u$) with the smallest label, encode the label $l_v$ of its adjacent vertex ($v$) as the next element of the code, and remove the vertex $u$ from 
\emph{Pen\_List}. If $v$ is now a pendant vertex, add it to \emph{Pen\_List}". Thus, in each iteration, an edge ($u,v$) is encoded. Although Pr{\"u}fer code 
comprises a string of vertex labels, each label $l_v$ in the code actually represents an edge ($u,v$). Hence, a Pr{\"u}fer code represents a \emph{sequence of edges} rather than vertices. Out of the set, \emph{Edge\_Seq}, of  all  possible edge-sequences in a tree, a Pr{\"u}fer encoding induces a \emph{sub-set}, 
\emph{Edge\_Seq\_Pr{\"u}fer}, of the set of edge-sequences. we next look at this sub-set, \emph{Edge\_Seq\_Pr{\"u}fer}.

For an edge to be selected in a particular   
iteration, one of its end-vertices should satisfy two conditions:
\begin{enumerate}
\item it should belong to \emph{Pen\_List},  and
\item it should have the smallest label among them.
\end{enumerate}  

Hence, in order to obtain a desired edge sequence, we need to take care of the above two conditions. The first condition is mostly affected by the structure of tree and hence, we have little control over it. However, the second condition depends on the labeling of vertices.  \emph{Edge\_Seq\_Pr{\"u}fer} is the subset of \emph{Edge\_Seq}, where an edge is prioritized on the basis of creating  a pendant vertex. An example to demonstrate this is given in Table \ref{table:edge_seq_eg}. Edges $e_1$ and $e_3$ of the tree can be named interchangeably because of the symmetry of the tree structure.  Note that, the size of the set  \emph{Edge\_Seq} is three, and  its elements are listed in the first column; first two sequences can be  encoded with Pr{\"u}fer code and so belong to \emph{Edge\_Seq\_Pr{\"u}fer}. The third sequence cannot be encoded by a Pr{\"u}fer code.

\begin{center}
 \setlength\extrarowheight{4pt}
\begin{table}[h]

\caption{Edge sequence example. }\label{table:edge_seq_eg}
\centering
\scriptsize

 \begin{tabular}{|c|c|c|} 
 \hline
 \multicolumn{3}{|c|}
 { \begin{tikzpicture}[baseline=8,shorten >=1pt]
   \tikzstyle{vertex}=[circle,fill=black!25,minimum size=12pt,inner sep=0pt]
 \node  at (5,0.5) {Example tree};
   \foreach \name/\x in {1/2, 2/4, 3/6, 4/8}
     \node[vertex] (G-\name) at (\x,0) {$ $};
     
 \foreach \name/\x in {e_1/3, e_2/5, e_3/7}
     \node  at (\x,-0.2) {$\name$};
     
   \foreach \from/\to in {1/2,2/3,3/4}
     \draw (G-\from) -- (G-\to);
   
 \end{tikzpicture}}\\
 \multicolumn{3}{|c|}{}\\
 \hline\hline
 
{Edge-Sequence}& Vertex labelling&Pr{\"u}fer code\\
\hline
$e_1-e_2-e_3$ & 
\begin{tikzpicture}[baseline=1,shorten >=1pt]
  \tikzstyle{vertex}=[circle,fill=black!25,minimum size=10pt,inner sep=0pt]

  \foreach \name/\x in {1/2, 2/3.5, 3/5, 4/6.5}
    \node[vertex] (G-\name) at (\x,0) {$\name$};
    
\foreach \name/\x in {e_1/2.8, e_2/4.3, e_3/5.8}
    \node  at (\x,-0.2) {$\name$};
    
  \foreach \from/\to in {1/2,2/3,3/4}
    \draw (G-\from) -- (G-\to);

\end{tikzpicture}& 2 3 \\
&&\\
$e_1-e_3-e_2$  & 
\begin{tikzpicture}[baseline=1,shorten >=1pt]
  \tikzstyle{vertex}=[circle,fill=black!25,minimum size=10pt,inner sep=0pt]

  \foreach \name/\x in {1/2, 3/3.5, 4/5, 2/6.5}
    \node[vertex] (G-\name) at (\x,0) {$\name$};
    
\foreach \name/\x in {e_1/2.8, e_2/4.3, e_3/5.8}
    \node  at (\x,-0.2) {$\name$};
    
  \foreach \from/\to in {1/3,3/4,4/2}
    \draw (G-\from) -- (G-\to);

\end{tikzpicture}& 3 4\\&&\\
$e_2-e_1-e_3$  & does not exists & Not possible\\
 \hline
\end{tabular}

\end{table}
\end{center}

\item \textit{The vertex-labels of $T$ determine the nature of the 
Pr{\"u}fer code.} 
\label{pruf_prop_2}

This is an important property and  follows from the above discussion of Property \ref{pruf_prop_1}. 

\item \textit{$T$ can be encoded to a set, $S_T$, of distinct Pr{\"u}fer codes.} 
\label{pruf_prop_3}

From Property \ref{pruf_prop_2}, every relabeling of $T$ furnishes a Pr{\"u}fer code unique to that labeling. Thus, given an unlabeled $T$, it can be represented by several Pr{\"u}fer codes that make up the set $S_T$. The number of such codes is determined by the number of unique labeling of the vertices of $T$.

\item \textit{Every  Pr{\"u}fer code in $S_T$ decodes to the tree structure of $T$.} 
\label{pruf_prop_4}

On applying the decoding algorithm on a Pr{\"u}fer code in $S_T$, the structure of $T$ is reconstructed.

\item \textit{Pr{\"u}fer sequences induce a partition on the set $S$, which is the set of Pr{\"u}fer codes of all possible labeled trees with $n$ vertices such that each partition represents a  specific tree structure.} 
\label{pruf_prop_5}

The Cayley's formula, $n^{n-2}$, gives the number of labeled trees with $n$ vertices \cite{PJCT95}. Pr{\"u}fer codes provide a bijective proof of Cayley's formula and so the size of $|S|$ is exactly equal to the number of labeled trees with $n$ vertices. For a given $n$, the number of unlabeled trees  is less than the size of $S$ because there exist several labeled trees for a given unlabeled tree.  From Property \ref{pruf_prop_3}, each unlabeled tree $T$, of size $n$ can be represented by a Pr{\"u}fer code in $S_T$. Also, from Property \ref{pruf_prop_4}, each of them uniquely reconstructs the  original tree $T$. 
Hence, they are mutually exclusive. Furthermore, since every Pr{\"u}fer code among the $n^{n-2}$ codes reconstructs to some tree of size $n$.  The union of the sets $S_T$ for all unlabeled trees is thus collectively exhaustive. 
Therefore, they induce a partition.

\end{enumerate}

\subsection{Encoding Methods}
\label{subsec:labelling_scheme}

From Property \ref{pruf_prop_3} and 
Property \ref{pruf_prop_4} it is clear that  we can select any code  from the partition $S_T$ (Property 
\ref{pruf_prop_5}) to represent the structure of a \emph{gtree}, $T_g$. Here we present three 
codes that are generated based on graph relabeling, label reordering, and tree 
relabeling. 

\begin{enumerate}[leftmargin=*]
\item \textbf{Direction-centric code (DCC).}
For directed acyclic graphs (DAG), a topological ordering of vertices preserves 
the direction of the edges. So, in this method, the vertices of a graph are 
labeled in a topological order. Each pair of vertices $(v_1,v_2)$, with the edge  
directed from $v_1$ to $v_2$, is labeled such that label of  $v_1$ is smaller 
than that of $v_2$.  Since combinatorial logic networks are represented as DAGs,  
DCC completely preserves the structure of logic networks.

\item \textbf{Path-centric code (PCC).}
\begin{figure}[b!]
\centering

\begin{subfigure}{0.5\textwidth}
\centering
\includegraphics[scale=0.65]{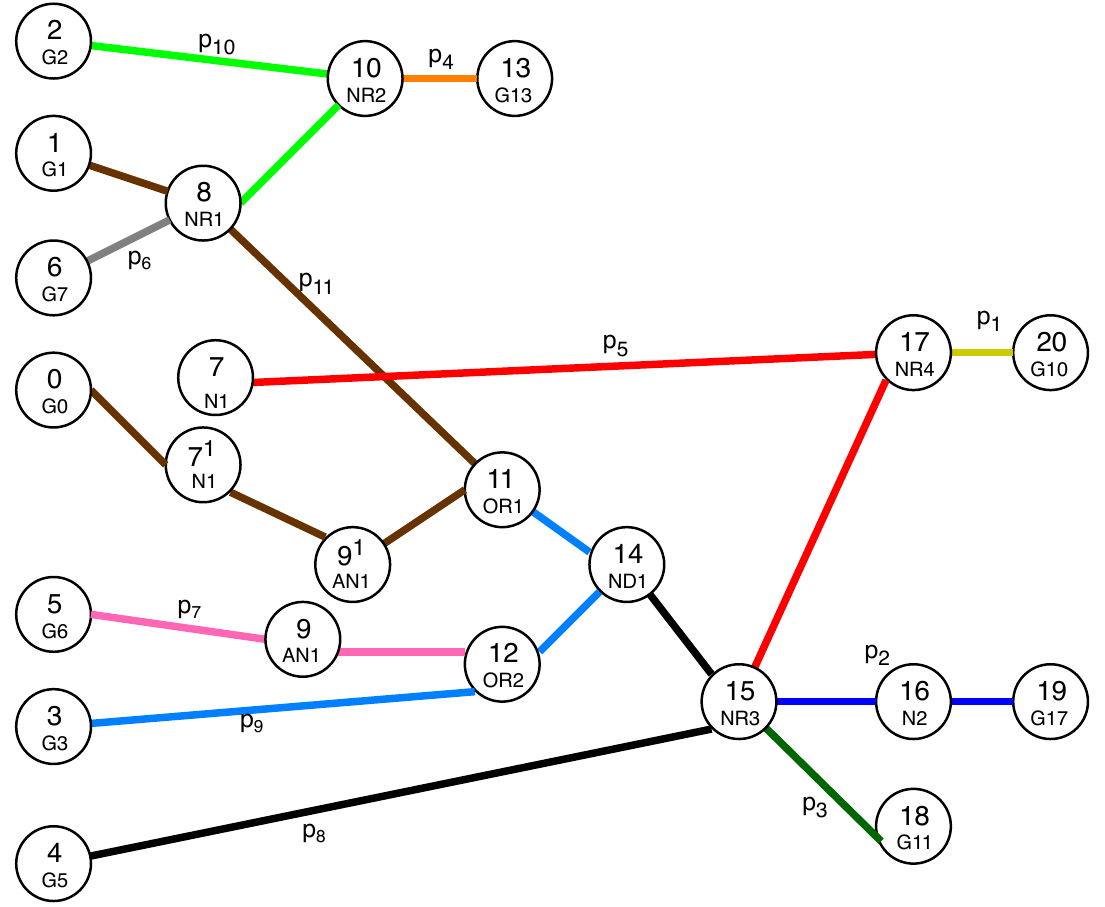}

\caption{Path partitions of the graph for circuit s27. }\hfill
\label{fig:tree_path_1}
\end{subfigure}

\begin{subfigure}{0.5\textwidth}
\hspace{1in}\includegraphics[scale=0.6]{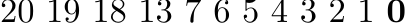}
\caption{Sequence of single pendant vertices derived from the above code}\hfill
\label{fig:tree_path_2}
\end{subfigure}

\begin{subfigure}{0.5\textwidth}
\centering
\includegraphics[scale=0.7]{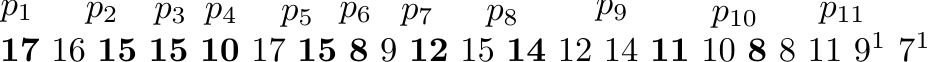}
\caption{Pr{\"u}fer code}\label{fig:tree_path_3}
\end{subfigure}
\caption{Example of PCC for \emph{g-tree} of s27.}\label{fig:tree_path}
\end{figure}
In PCC, the labeling scheme is same as DCC, so it preserves the direction of the edges. While in DCC, the edges are not directly interpretable, in PCC, we make the edges interpretable by introducing a change in the ordering of vertices 
keeping their labels intact. Up till now we have considered that the vertex-labels also determine 
their ordering, smaller label meaning higher order. In case of PCC, we introcduce a smal change in the ordering for for the leaf nodes of $T_g$. Let the set of leaf nodes of $T_g$ be $V^T_{leaf}$. Here, we assume that the order of vertices in $V^T_{leaf}$ is higher than those of non-leaf vertices ($V^T \setminus V^T_{leaf}$). When this condition in ordering is introduced, the vertices in the set  $V^T \setminus V^T_{leaf}$ that appear in  \emph{Pen\_List} are processed before those in $V^T_{leaf}$ 
during encoding/decoding (Refer Section \ref{subsec:pruf_prop} under Property \ref{pruf_prop_1}). This has the following implication:  For a leaf vertex  $u_{leaf}$ of $T_g$ we define two terms: (i) \emph{junction point} ($u_m$),  the first vertex, of degree greater than two, reachable from  $u_{leaf}$, (ii) \emph{path-vertices}, the set of two-degree vertices that lie between $u_{leaf}$ and its junction point.   In the Pr{\"u}fer code, the path-vertices  appear consecutively, in the same sequence as in the path, $u_{leaf} \Leadsto u_m$, followed by the junction vertex $u_m$. Thus, each of the leaf node, except the last one, induces a path. Also, the order of paths follows that of the corresponding leaf-nodes. 

The edges can be easily reconstructed from their labels if we mark the vertices corresponding to the junction points. They can be marked by
traversing the code from right-to-left, and marking those vertex labels which have already appeared (Observation \ref{ob:vertex_in_code}).

An example of PCC for circuit s27 is given in Figure \ref{fig:tree_path}. The tree along with eleven paths is shown in Figure  \ref{fig:tree_path_1}. Note that the vertices in  $V^T_{leaf}$ are considered in reverse order; higher label means higher order (Figure  \ref{fig:tree_path_2}). The Pr{\"u}fer code with marked junction points is shown in Figure  \ref{fig:tree_path_3}. For example, the leaf-node 3 induces Path $p_9$, whose path-vertices \{12,14\} and vertex at junction point, vertex  labeled 11 appear sequentially in the code.

\item \textbf{Leaf-centric code (LCC).} 
Although the edges can be reconstructed directly from PCC, the leaf-node needs to be computed beforehand.  LCC is a fully edge-interpretable code. The index of a label gives the label of its adjacent vertex. This is possible because each vertex-label in the code represents an edge (Property \ref{pruf_prop_1}).  
This is accomplished by complete relabeling of  $T_g$, denoted by $T_{lcc}^1$,
which is done as follows: The nodes are partitioned into sets called \emph{leaf-stage}. The leaves of $T_g$ are assigned to the set representing the first stage, $LS1$. Next, on removing the nodes in  leaf-stage-1, the leaves of the new tree $T_{lcc}^2$ are assigned to $LS2$. Iteratively, the partitions are formed until the entire tree is processed. The final leaf-stage, $LSk$,  consists of either a node or a pair of nodes, and is called the \emph{tree center} \cite{harary71}. The labeling is done iteratively as follows. Vertices of  $LS1$ are labeled from 1 to $|LS1|$. Iteratively, the vertices in $k^{th}$  set is labeled $\sum^{k}_1 |LS(i-1)| + 1 $ to  $\sum^{k}_1 |LSi|$. Consider a set $REP \subset V^T$, where the vertices in $REP$ were formed by splitting a particular vertex in $G$. Let their labels be $l_{1} \cdots l_{m}$. Let $v \in REP$ have the smallest label ($l_m'$).  For each vertex $u$ in $REP$, with label $l_u$, their label is modified as  $l_{u}^{l_m'}$. The additional index is used to store the replica information and does not affect the ordering of vertices while encoding. Such labeling strategy will enforce the label and its index to preserve an edge. Also, the edges belonging to a leaf-stage appear consecutively in LCC.

There are scope for further improving the interpretability of LCC by adjusting the labels within each leaf-stage. The labels in a leaf-stage subsequence can be made to appear in an ascending order. Considering the tree to be rooted at its center, the vertices can be assigned to different levels progressively. The labeling starts from the vertex (vertices) in the root (level 0). The first vertex is labeled $n$ and subsequent labeling is processed in a decremental order. The labeling is done while moving from lower-level to higher-level nodes, while prioritizing them based on the leaf-stage (higher leaf-stage first). The vertices within a level and belonging to the same leaf-stage are prioritized based on the label of their parents (vertex with higher-labeled parent first).

Lastly, the direction of the edges can be encoded by marking the labels which represent the edges that are directed one way (converging or diverging) and leaving the edges in the opposite direction unmarked. Thus, besides DAG any directed  graph can also be encoded.

\begin{figure}[h]
\centering
\begin{subfigure}{0.5\textwidth}
\includegraphics[trim=0 40 0 0,clip,scale=0.55]{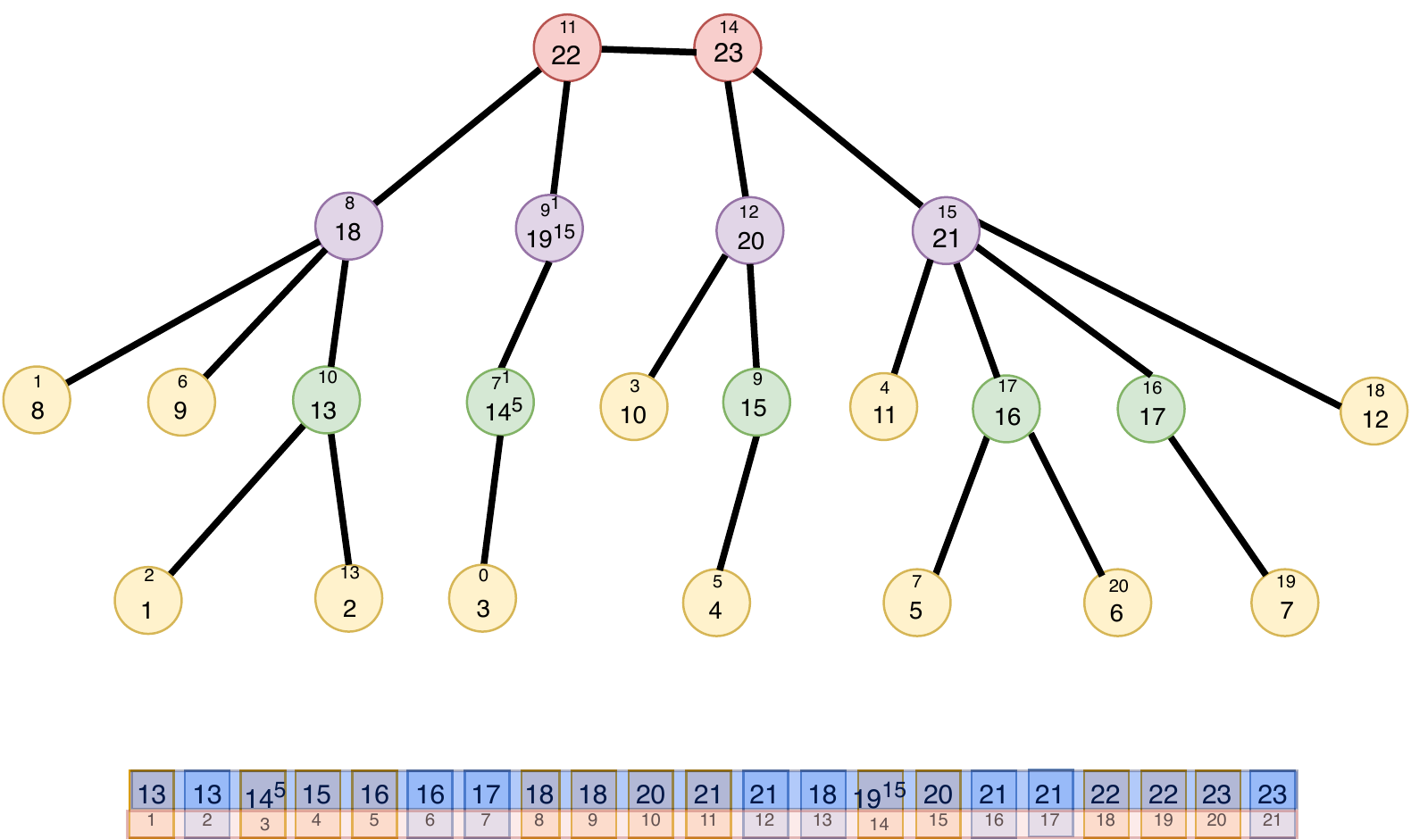}

\caption{Relabeled tree of s27 for LCC. The original label is shown in small font. Leaf-stage of vertices is color coded. Vertices with the same level appear  horizontally.}\hfill
\label{fig:lcc_1}
\end{subfigure}

\begin{subfigure}{0.5\textwidth}

\includegraphics[trim=20 0 37 220,clip,scale=0.55]{figures/s27/LCC/lcc.pdf}
\caption{LCC and the index}\label{fig:lcc_2}
\end{subfigure}
\caption{Example of LCC for \emph{g-tree} of s27.}\label{fig:lcc}
\end{figure}

\end{enumerate}

An example of LCC for s27 is given in Figure \ref{fig:lcc}. Figure \ref{fig:lcc_1} shows the tree where the root is the edge with vertices marked in red. The new labels are shown in the Figure \ref{fig:lcc_1}. Also, the labeling of vertices within each stage is taken care of so as to improve interpretability. This is reflected in the LCC  given in Figure \ref{fig:lcc_2}. Here, the label of the vertex with the corresponding edge incident on it, is marked in orange and one with corresponding edge divergent from it is marked in blue.
\subsection{Learnable Representation}
\label{subsec:graph_embedding}

Data samples are commonly represented as vectors not only in statistical inference but also in ML, where they are called feature vector. Image data can be viewed as vectors since they are regularly structured on a rectangular lattice.  Such representation offers computational ease and lends strong mathematical foundation \cite{FMSPC12} where huge repository of operations have been defined for vectors. Thus, there are numerous algorithms available that can be used for analyzing vector data in ML.  Graphs, however, comprise highly unstructured data and hence cannot naturally be represented as vectors. Although an adjacency matrix can be viewed as structured data, it has several issues which render it difficult to be used  as feature vectors. Firstly, for large graphs, the size ($|V|^2$) is too large for implementation. Secondly, in scenarios such as graph classification where a single graph is a data sample, it required that the graphs be of similar size. Lastly, even if they are of same size, we need a method to order the nodes so as to induce some correspondence among the nodes of different sample graphs.

Graph embedding has presently emerged as a common solution to above mentioned problems. It captures a certain properties of the graph in the form of vectors. There are of two kinds: (i) node embedding, where each node is represented by a vector. They are used mostly for node classification and link prediction of a graph. They are mostly dependent on the neighborhood information and  first/second degree proximity. Some of the examples are Node2Vec \cite{GKDD16}, LINE \cite{LINE15}, DeepWalk \cite{PKDD14}; (ii) whole graph embedding. Here the entire graph is represented by a vector which captures some of its properties. They are mostly applicable for graph classification. Such related work includes graph kernels \cite{YVSIGKDD15} and Subgraph2Vec \cite{AMCorr16}.

Since the aim of our representation is to capture the structural properties of the entire circuit-graph, it needs  whole-graph type embedding. However, the existing methods capture only abstracted properties of graph. For example, graphlet-based kernels represent the counts of different kinds of graphlets as a vector. More than just graph classification, we aim to learn structural features from the graph. The proposed Pr{\"u}fer code captures the structure of the entire graph, and it can handle the scalability issue mentioned earlier. However, the other two issues need to be addressed to make it suitable for graph embedding. The second issue can be handled if the assumption on the size of the graphs (the number of edges in this case), holds, i.e., they are of same order. For graphs representing logical circuits, we can synthesize similar-sized circuits to form the training data. Any difference in their sizes can be handled augmenting them with  pseudo-vertices/edges as in \cite{MMICML16}. For node correspondence, ordering of the nodes can be accomplished using some special properties \cite{MMICML16} such as node-degree. In case of circuit-graphs, some circuit properties such the level of a logic gate in the circuit can also be used.

Once we obtain an ordering of the vertices of the graph $G(V,E)$, we can follow this ordering while traversing the graph during SCESOR traversal. This will enhance th possibility of obtaining a \emph{gtree} thus will be a unique tree of the graph based on vertex ordering. We can incorporate such ordering to the encoding schemes dicussed in the previous section. Since the topological ordering of a DAG is not unique,  in the case of DCC or PCC, the  vertices that have the same level are not ordered. So, in such vertices in the graph can be labeled following  the ordering of nodes. Also, in the case of LCC, those vertices, which belong to the same tree-level, leaf-stage, or have the same parent are not ordered. So, these additional ordering properties can be applied here also. 

\section{Conclusion and Future Work}
\label{sec:conclusion_ch5}
In this work, we have demonstrated a proof-of-concept 
for lossless and compact encoding
of large graphs using just a linear-size sequence of vertex labels. In order to encode graphs with Pr{\"u}fer codes, we have proposed a method called $\mathcal{GT}$-enhancement, to represent a graph by a tree. We have proposed two techniques for $\mathcal{GT}$-enhancement, and the second technique called SCESOR,  allows such tree to be represented by a single code.  The focus of the encoding is to capture the structural property of a  graph representing a digital circuit, which is essentially a DAG. Among various graph-representation methods used for machine-learning framework, a major issue is to ensure lossless encoding of the structure of the graph. The method based on Pr{\"u}fer codes completely overcomes this issue. Moreover, we have proposed a labeling technique that preserves the direction of the edges and improves their interpretability. We have discussed the learnability of the code and how they can be used for graph embedding.  In general, a learnable representation of the code can potentially be applied for geometric deep learning. Additionally Pr{\"u}fer code offers a 1-D representation of the graph, which is often a requirement for geometric deep learning \cite{MLSPM17}. Besides, in the case of  \emph{g-trees} generated from graphs with ordered vertices, the interpretable codes such as LCC bring certain structuredness in the representation, which is required for such applications. Use of such codes in deep learning framework like RNN or CNN can be explored in the future.

{
\tiny
\bibliographystyle{IEEEtran}
\bibliography{references/deep-learning-references,references/encoding_references,references/dts_references,references/my,references/ml_vlsi,references/references_X,references/ml_vlsi_references}
}

\end{document}

%% file: tex_files/tree_part_table.tex
\flushbottom
\begin{table*}[t]
\caption{Results on  logic circuits in ISCAS'89 and ITC'99 benchmark-suites. }\label{table:pruf_experimental_result}
\label{table:experimental_result}
\centering
\scriptsize
 \begin{tabular}{|c|c|c|c|c|c|c|c|c|c|} 
 \hline
Circuit 	&	\#vertices	&	\#edges	&Pr{\"u}fer code&	\#extra\_	&	\#be\_	&	 \#edge\_	&	\#vertex	&	\#label\_	&	CPU-time	\\
&&&length&labels ($|L|$)&trees&swap&\_split&swap&in sec.\\
 \hline
 \multicolumn{9}{|c|}{ISCAS'89 benchmark circuits \cite{iscas89}}\\
 \hline
s5378	&	3206	&	4435&	4434	&	27	&	2	&	1	&	1230	&	606	&	7.76	\\
s9234	&	6094	&	8235&	8234	&	292	&	4	&	25	&	2142	&	1048	&	15.29	\\
s13207	&	9441	&	12048	&12047	&	450	&	4	&	14	&	2608	&	1184	&	24.56	\\
s15850	&	11067	&	14380	&14379	&	568	&	4	&	10	&	3314	&	1594	&	28.74	\\
s38417	&	25585	&	33969	&33968	&	1202	&	4	&	4	&	8385	&	4535	&	77.78	\\
s38584	&	22447	&	34497&	34496	&	2325	&	4	&	5	&	12051	&	5628	&	125.77	\\
\hline
 \multicolumn{9}{|c|}{ITC'99 benchmark circuits \cite{itc99}}\\
 \hline
b13s	&	392	&	601	&600	&	0	&	2	&	7	&	210	&	111	&	1.12	\\
b14s	&	5020	&	9862&	9861	&	1324	&	3	&	26	&	4843	&	1881	&	40.88	\\
b15s	&	9343	&	19068&	19067	&	2534	&	3	&	39	&	9726	&	3909	&	87.38	\\
b17s	&	25615	&	52447&	52446	&	7282	&	3	&	42	&	26833	&	10175	&	307.35	\\
b20s	&	9909	&	19555&	19554	&	2739	&	3	&	21	&	9647	&	3620	&	98.39	\\
b21s	&	10293	&	20300&	20299	&	2387	&	3	&	46	&	10008	&	3871	&	104.73	\\
b22s	&	15836	&	31304	& 31303	&	4105	&	3	&	11	&	15469	&	5856	&	158.33	\\
 \hline
\end{tabular}

\end{table*}
\flushbottom

%% file: graph_representation.bbl
\begin{thebibliography}{10}
\providecommand{\url}[1]{#1}
\csname url@samestyle\endcsname
\providecommand{\newblock}{\relax}
\providecommand{\bibinfo}[2]{#2}
\providecommand{\BIBentrySTDinterwordspacing}{\spaceskip=0pt\relax}
\providecommand{\BIBentryALTinterwordstretchfactor}{4}
\providecommand{\BIBentryALTinterwordspacing}{\spaceskip=\fontdimen2\font plus
\BIBentryALTinterwordstretchfactor\fontdimen3\font minus
  \fontdimen4\font\relax}
\providecommand{\BIBforeignlanguage}[2]{{%
\expandafter\ifx\csname l@#1\endcsname\relax
\typeout{** WARNING: IEEEtran.bst: No hyphenation pattern has been}%
\typeout{** loaded for the language `#1'. Using the pattern for}%
\typeout{** the default language instead.}%
\else
\language=\csname l@#1\endcsname
\fi
#2}}
\providecommand{\BIBdecl}{\relax}
\BIBdecl

\bibitem{ISARX20}
I.~Chami, S.~Abu-El-Haija, B.~Perozzi, C.~Ré, and K.~Murphy, ``Machine
  learning on graphs: A model and comprehensive taxonomy,'' 2020.

\bibitem{HVTKDE18}
H.~Cai, V.~W. Zheng, and K.~C. Chang, ``A comprehensive survey of graph
  embedding: Problems, techniques, and applications,'' \emph{IEEE Transactions
  on Knowledge and Data Engineering}, vol.~30, no.~9, pp. 1616--1637, Sept
  2018.

\bibitem{FMSPC12}
Y.~Fu and Y.~Ma, \emph{Graph Embedding for Pattern Analysis}.\hskip 1em plus
  0.5em minus 0.4em\relax New York: Springer Publishing Company, Incorporated,
  2012.

\bibitem{LWCorr17}
W.~L. Hamilton, R.~Ying, and J.~Leskovec, ``Representation learning on graphs:
  Methods and applications,'' \emph{CoRR}, vol. abs/1709.05584, 2017.

\bibitem{MLSPM17}
M.~M. Bronstein, J.~Bruna, Y.~LeCun, A.~Szlam, and P.~Vandergheynst,
  ``Geometric deep learning: Going beyond {E}uclidean data,'' \emph{IEEE Signal
  Processing Magazine}, vol.~34, no.~4, pp. 18--42, July 2017.

\bibitem{GKDD16}
A.~Grover and J.~Leskovec, ``Node2vec: Scalable feature learning for
  networks,'' in \emph{Proc. KDD}.\hskip 1em plus 0.5em minus 0.4em\relax ACM,
  2016, pp. 855--864.

\bibitem{LINE15}
J.~Tang, M.~Qu, M.~Wang, M.~Zhang, J.~Yan, and Q.~Mei, ``{LINE:} large-scale
  information network embedding,'' \emph{CoRR}, vol. abs/1503.03578, 2015.

\bibitem{DBNIPS16}
M.~Defferrard, X.~Bresson, and P.~Vandergheynst, ``Convolutional neural
  networks on graphs with fast localized spectral filtering,'' in \emph{Proc.
  NIPS}.\hskip 1em plus 0.5em minus 0.4em\relax Curran Associates, Inc., 2016,
  pp. 3844--3852.

\bibitem{MMICML16}
M.~Niepert, M.~Ahmed, and K.~Kutzkov, ``Learning convolutional neural networks
  for graphs,'' in \emph{Proc. ICML}, 2016, pp. 2014--2023.

\bibitem{FMTNN09}
F.~{Scarselli}, M.~{Gori}, A.~C. {Tsoi}, M.~{Hagenbuchner}, and
  G.~{Monfardini}, ``The graph neural network model,'' \emph{IEEE Trans. NN},
  vol.~20, no.~1, pp. 61--80, Jan 2009.

\bibitem{GWCoRR17}
G.~Wang, ``A novel neural network model specified for representing logical
  relations,'' \emph{CoRR}, vol. abs/1708.00580, 2017.

\bibitem{WEISCAS18}
W.~Haaswijk, E.~Collins, B.~Seguin, M.~Soeken, F.~Kaplan, S.~Süsstrunk, and
  G.~D. Micheli, ``Deep learning for logic optimization algorithms,'' in
  \emph{Proc ISCAS}, 2018, pp. 1--4.

\bibitem{YRHOST17}
Y.~Dai and R.~K. Braytont, ``Circuit recognition with deep learning,'' in
  \emph{Proc. HOST}, 2017, pp. 162--162.

\bibitem{YHDAC19}
Y.~Ma, H.~Ren, B.~Khailany, H.~Sikka, L.~Luo, K.~Natarajan, and B.~Yu, ``High
  performance graph convolutional networks with applications in testability
  analysis,'' in \emph{Proc. {DAC}}, Jun 2019, pp. 1--6.

\bibitem{MBWIRES20}
M.~Pradhan and B.~B. Bhattacharya, ``A survey of digital circuit testing in the
  light of machine learning,'' \emph{WIREs Data Mining and Knowledge
  Discovery}, p. e1360, 2020.

\bibitem{MBTCAD18}
M.~{Pradhan}, B.~B. {Bhattacharya}, K.~{Chakrabarty}, and B.~B. {Bhattacharya},
  ``Predicting ${X}$ -sensitivity of circuit-inputs on test-coverage: A
  machine-learning approach,'' \emph{Proc. TCAD}, vol.~38, no.~12, pp.
  2343--2356, 2019.

\bibitem{PAMP18}
H.~Prufer, ``Neuer beweis eines satzes uber permutationen,'' \emph{Arch. Math.
  Phys.}, vol.~27, pp. 742--744, 1918.

\bibitem{iscas89}
F.~Brglez, D.~Bryan, and K.~Kozminski, ``Combinational profiles of sequential
  benchmark circuits,'' in \emph{Proc. ISCAS}, 1989, pp. 1929--1934.

\bibitem{itc99}
F.~Corno, M.~Reorda, and G.~Squillero, ``{RT}-level {ITC}'99 benchmarks and
  first {ATPG} results,'' \emph{IEEE Design and Test of Computers}, vol.~17,
  no.~3, pp. 44--53, Jul 2000.

\bibitem{BBGD07}
T.~Biedl and F.~J. Brandenburg, ``Partitions of graphs into trees,'' in
  \emph{Proc. Graph Drawing}, 2007, pp. 430--439.

\bibitem{JJGECCO01}
J.~Gottlieb, B.~A. Julstrom, G.~R. Raidl, and F.~Rothlauf, ``Pr\"{u}fer
  numbers: A poor representation of spanning trees for evolutionary search,''
  in \emph{Proc. GECCO}, 2001, pp. 343--350.

\bibitem{WWJSEA09}
X.~Wang, L.~Wang, and Y.~Wu, ``An optimal algorithm for prufer codes,''
  \emph{JSEA}, vol.~2, pp. 111--115, 01 2009.

\bibitem{Cormen}
T.~H. Cormen, C.~Stein, R.~L. Rivest, and C.~E. Leiserson, \emph{Introduction
  to Algorithms}, 2nd~ed.\hskip 1em plus 0.5em minus 0.4em\relax Cambridge:
  McGraw-Hill Higher Education, 2001.

\bibitem{KACM62}
A.~B. Kahn, ``Topological sorting of large networks,'' \emph{Commun. ACM},
  vol.~5, no.~11, pp. 558--562, Nov. 1962.

\bibitem{PJCT95}
P.~W. Shor, ``A new proof of cayley's formula for counting labeled trees,''
  \emph{Journal of Combinatorial Theory, Series A}, vol.~71, no.~1, pp. 154 --
  158, 1995.

\bibitem{harary71}
F.~Harary, \emph{Graph Theory}, ser. Addison Wesley {S}eries in
  {M}athematics.\hskip 1em plus 0.5em minus 0.4em\relax Boston: Addison-Wesley,
  1971.

\bibitem{PKDD14}
B.~Perozzi, R.~Al-Rfou, and S.~Skiena, ``Deepwalk: Online learning of social
  representations,'' in \emph{Proc. KDD}.\hskip 1em plus 0.5em minus
  0.4em\relax ACM, 2014, pp. 701--710.

\bibitem{YVSIGKDD15}
P.~Yanardag and S.~Vishwanathan, ``Deep graph kernels,'' in \emph{Proc. ACM
  SIGKDD}.\hskip 1em plus 0.5em minus 0.4em\relax New York, NY, USA: ACM, 2015,
  pp. 1365--1374.

\bibitem{AMCorr16}
A.~Narayanan, M.~Chandramohan, L.~Chen, Y.~Liu, and S.~Saminathan,
  ``subgraph2vec: Learning distributed representations of rooted sub-graphs
  from large graphs,'' \emph{CoRR}, vol. abs/1606.08928, 2016.

\end{thebibliography}
